\documentclass[sigconf]{acmart}

\usepackage{booktabs} 

\usepackage{textcomp}
\usepackage{latexsym}
\usepackage{graphicx} 
\usepackage{microtype}
\usepackage{pifont}
\usepackage[skip=0pt,font=small]{caption}
\usepackage[]{subcaption}
\usepackage{verbatim}
\usepackage{framed}
\usepackage{color, colortbl}
\usepackage[colorinlistoftodos]{todonotes}
\usepackage{amssymb}
\usepackage{amsmath}
\usepackage{breqn}
\usepackage{capt-of}
\usepackage{multirow}
\usepackage{algorithmicx}
\usepackage{algorithm}
\usepackage{algcompatible}
\usepackage{paralist}
\usepackage{enumitem}
\usepackage{balance}
\usepackage[all]{nowidow}

\definecolor{high}{HTML}{FFCE8E}
\DeclareMathOperator*{\argmax}{arg\,max}

\setcopyright{rightsretained}

\acmDOI{}

\acmISBN{}

\acmConference[]{}{}{} 
\acmYear{}
\copyrightyear{2018}

\acmPrice{}

\acmSubmissionID{}

\begin{document}
\title{TFW, DamnGina, Juvie, and Hotsie-Totsie: On the Linguistic and Social Aspects of Internet Slang}

\author{Vivek Kulkarni}
\orcid{}
\affiliation{%
  \institution{Department of Computer Science \\ 
  	University of California, Santa Barbara}
  \city{Santa Barbara} 
  \state{California} 
  \postcode{43017-6221}
}
\email{vvkulkarni@ucsb.edu}

\author{William Yang Wang}
\orcid{}
\affiliation{%
	\institution{Department of Computer Science \\ 
		University of California, Santa Barbara}
	\city{Santa Barbara} 
	\state{California} 
	\postcode{43017-6221}
}
\email{william@cs.ucsb.edu}

\begin{abstract}
Slang is ubiquitous on the Internet. 
The emergence of new social contexts like micro-blogs, question-answering forums, and social networks has enabled slang and non-standard expressions to abound on the web. 
Despite this, slang has been traditionally viewed as a form of non-standard language -- a form of language that is not the focus of linguistic analysis and has largely been neglected. 

In this work,  we use \textsc{UrbanDictionary} to conduct the first large scale linguistic analysis of slang and its social aspects on the Internet to yield insights into this variety of language that is increasingly used all over the world online.
First,  we begin by computationally analyzing the phonological, morphological and syntactic properties of slang in general. We then study linguistic patterns in four specific categories of slang namely \emph{alphabetisms, blends, clippings, and reduplicatives}. 
Our analysis reveals that slang demonstrates extra-grammatical rules of phonological and morphological formation that markedly distinguish it from the standard form shedding insight into its generative patterns. 
Second, we follow up by analyzing the social aspects of slang where we study subject restriction and stereotyping in slang usage. 
Analyzing tens of thousands of such slang words reveals that the majority of slang on the Internet belongs to two major categories: \emph{sex} and \emph{drugs}. 
Further analysis reveals that not only does slang demonstrate prevalent gender and religious stereotypes but also an increased bias where even names of persons are associated with higher sexual prejudice than one might encounter in the standard form.

In summary, our work suggests that slang exhibits linguistic properties and lexical innovation that strikingly distinguish it from the standard variety. Moreover, we note that not only is slang usage not immune to prevalent social biases and prejudices but also reflects such biases and stereotypes more intensely than the standard variety.
\end{abstract}

%



\maketitle

\emergencystretch 3em
\section{Introduction}
\label{sec:intro}
\begin{figure}[t!]
	\includegraphics[scale=0.5, width=\columnwidth]{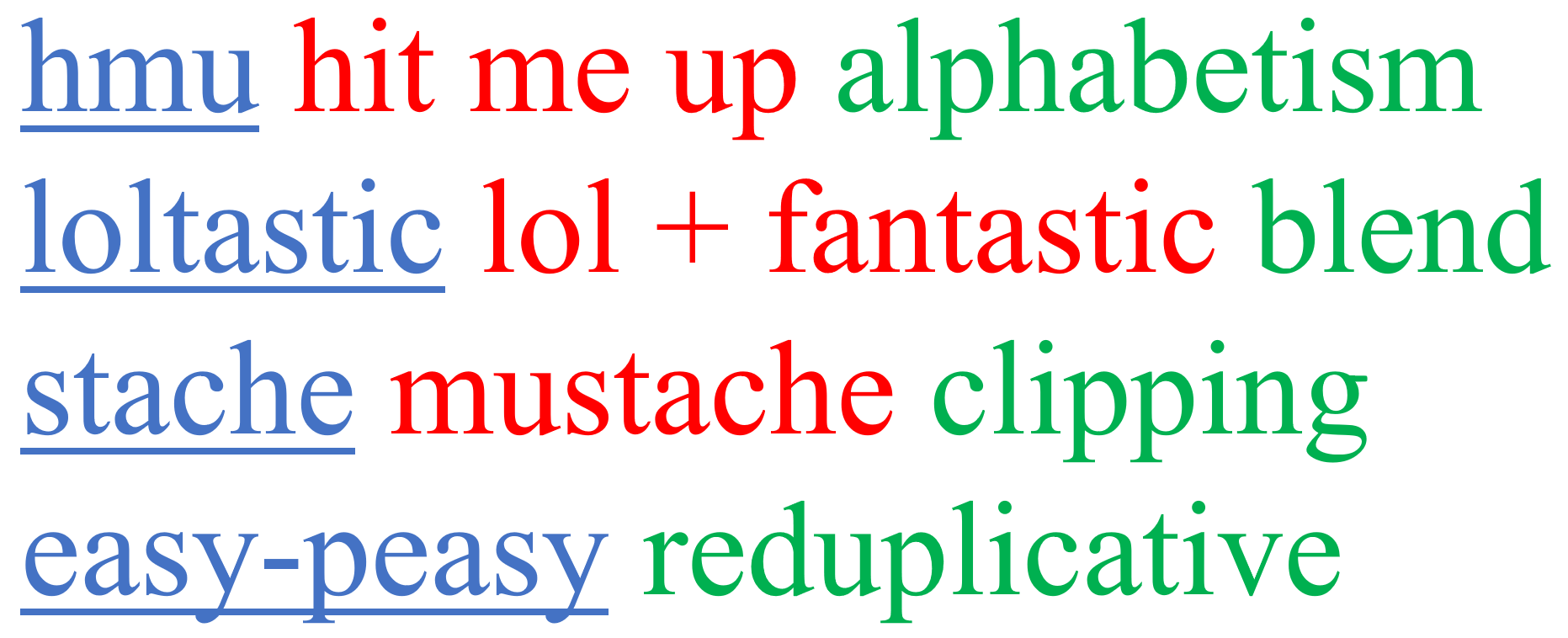}
	\caption{Sample words depicting some of the linguistic features in slang on the Internet. Note the presence of simple alphabetisms as well as more complex phenomena like blending (\texttt{loltastic}).}
	\label{fig:crown_jewel}
\end{figure}
The Internet is global, diverse, and dynamic, all of which are reflected in its language. 
One aspect of this diversity and dynamism is the abundance of slang and non-standard varieties -- an aspect which has traditionally received little linguistic attention. In fact \citet{labov1972some} argues that all articles focusing on slang should be assigned to an ``extra-linguistic darkness''. According to \citet{ebleslang},  while the development of socio-linguistics has legitimized the study of slang, slang does not naturally fit into the controlled framework of socio-linguistics where correlations between social factors like age, gender and ethnicity with language use are studied, and is better explained through the lens of social connections (like personal kinship). In other words, slang is firmly grounded in social connections and contexts  enabling ``group identity''. 
However, the evolution of Internet and social media has radically transformed these social contexts \cite{ebleslang,Crystal:2011:ILS:2011923}. First, slang is no longer restricted to oral communication but is now widely prevalent on the Internet in written form. Second, the notion of a \emph{group} associated with slang now extends to a \emph{network} that is increasingly global where members are not necessarily associated by friendship or face-to-face interactions but are participants in new and evolving social contexts like forums and micro-blogs \cite{ebleslang,Crystal:2011:ILS:2011923}.
Consequently, \citet{ebleslang} claims \emph{``slang is now world-wide the vocabulary of choice of young people (who compose the majority of inhabitants of the earth) and reflects their tastes in music, art, clothing and leisure time pursuits''} and further argues that its study has fallen behind since it is not an integral part of socio-linguistics. 

Here, we address this gap and conduct the first large scale computational analysis of slang on the Internet using \textsc{UrbanDictionary}, addressing both linguistic as well as social aspects.

First, we characterize linguistic patterns dominant in the formation of slang and provide supporting evidence of its distinctiveness from the standard variety not only supporting claims made by \citet{mattiello2008introduction} but also revealing new insights into the phonological, morphological and syntactic patterns evident in slang through a large scale quantitative analysis.
Figure \ref{fig:crown_jewel} illustrates some of these patterns. 
\texttt{hmu} is an \textsc{Alphabetism}, more specifically an \textsc{Initialism} for \texttt{hit me up}. 
\textsc{Blends} are formed by mixing parts of existing words. \footnote{These words are also called portmanteaus.} 
For example, \texttt{loltastic} is a blend of \texttt{lol} and \texttt{fantastic}. 
Other examples include words like \texttt{netizen, infotainment, frenemy, and bootylicious}.
While \textsc{Blends} are formed by the mixture of two or more words, \textsc{Clippings} are formed by \emph{shortening} the original word. \texttt{stache} is an example since it is a shortened form of \texttt{mustache}. 
Finally, \textsc{Reduplicatives}, also called \emph{echo words} consist of word pairs, where the first word is phonetically similar to the second word. Examples are \texttt{teenie-weenie, artsy-fartsy, boo-boo, chick-flick}.
While these patterns are not exhaustive, they suggest that slang exhibits rich and varied linguistic formation patterns.
We therefore, conduct an in-depth investigation into the phonological, morphological and syntactic properties of slang words and contrast them with the standard form yielding insights into the linguistic mechanisms at play in slang formation (see \textbf{Section \ref{sec:linguistic}}). 

Second, we analyze the social aspects associated with slang. Inline with \citet{ebleslang} who claims \emph{``One of the greatest challenges to scholarship in slang is to fit slang into the current conversations going on in sociolinguistics
about such topics as identity, power, community formation, stereotyping, discrimination and the like''}, we investigate two social aspects of slang: (a) \textbf{Subject Restriction} -- Slang is strongly associated with certain subjects like \textsc{Sex}, \textsc{Drugs} or \textsc{Food}. We study this aspect by proposing a model to classify slang words into set of $10$ pre-defined categories to reveal dominant subjects (b) \textbf{Stereotyping} -- We analyze and quantify biases and stereotypes evident in slang usage and show evidence of gender stereotypes, sexual and religious prejudices -- prejudices that are much more extreme in slang than those observed in the standard form (see \textbf{Section \ref{sec:social}}).          

In a nutshell therefore, our contributions are:
\begin{enumerate}
	\item \textbf{Linguistic Aspects of Slang}: We analyze the phonological, morphological and syntactical properties of slang and contrast them  with those found in the standard form revealing insights into patterns governing slang formation. 
	\item \textbf{Social Aspects of Slang}: We analyze two social aspects associated with slang: (a) \emph{Subject Restriction} and (b) \emph{Stereotypes and prejudices} reflected in slang usage online.
\end{enumerate}
Altogether, our results shed light on both the linguistic and social aspects governing slang formation and usage on the Internet.

\section{Datasets}
\paragraph{Slang Data Set} In discussions of slang, one controversial issue has always been its definition. 
\citet{mattiello2008introduction} notes that there are multiple views to characterize slang. 
One dominant view (a sociological view) adopted by \cite{eble2012slang,munro1989slang} is that it primarily enables one to identify with a specific group/cohort. 
Others like \cite{quirk2010comprehensive} adopt a more stylistic view and are of the opinion that slang should be categorized according to attitudes that vary from  ``the casual to the vulgar''. 
Still others emphasize the novelty of slang and characterize slang as a variety of language that is inclined towards lexical innovation \cite{mencken1967american,dumas1978slang,sornig1981lexical}. 
In our work, we adopt a broad definition of slang which includes non-standard expressions (words) and emphasizes both the sociological viewpoint as well as the viewpoint of lexical innovation and novelty.
We constructed a large data set of  slang, by scraping the popular online slang dictionary \textsc{UrbanDictionary} as of July 2017. 
For each word, we obtain the top $10$ definitions (when available) as well as their example usage and vote counts. 
We remove very rare words and short-lived trends by considering only instances with at-least $100$ votes (weeding out rare/noisy slang like \texttt{jeff cohoon} which has no votes or short-lived trends such as \texttt{Naimbia} which was added after President Trump mistook it for a country). 
This yields a dataset of $128,807$ words and $328,170$ definitions spanning the time period $1999-2017$ leaning towards relatively long-lived slang words rather than short-lived fads/phenomena. 
We note that even though \textsc{Urban Dictionary} was created in $1999$, the majority of the words entries are introduced after $2005$ 
which interestingly coincides with evolution of social media like \textsc{Facebook} and \textsc{Twitter}.   
Finally, we observed that the fraction of single word slang is slightly higher than $50\%$, suggesting that a significant fraction of slang consists of multiple words or phrases in contrast with standard English where almost all dictionary entries are exclusively single words or two-word phrases. 

\paragraph{Standard English (SE)} To contrast slang with the more standard usage of English, we consider a list of $67,713$ words released by the \textsc{Dictionary Challenge} \cite{hill2015learning} which consists of words and their definitions obtained from electronic resources like \emph{Webster's}, \emph{WordNet}, \emph{The Collaborative International Dictionary of English} and \emph{Wiktionary} and pre-dominantly contains words in standard usage (including technical jargon, scientific terms etc). 
\section{Linguistic Aspects of Slang}
\label{sec:linguistic}
Having described our datasets, we now proceed to analyze several linguistic aspects of slang words and contrast them with those in Standard English (SE).
\subsection{Phonology}
\citet{mattiello2005pervasiveness} suggests  that slang incorporates broad phonological phenomena like echoisms, mis-pronunciations and assimilation. However little is known about specific phonological patterns/properties of slang where lexical innovation is so rampant. 
\emph{What are the phonological properties of slang and how do they differ from those of words in Standard English?}

To analyze phonological properties, we obtain the phoneme representation of each word using \texttt{G2PSeq2Seq} \cite{yao2015sequence}, a neural pre-trained model  for the task of grapheme (letter) to phoneme conversion trained on the \emph{CMU Pronouncing Dictionary}.
To illustrate, the phoneme representation for \texttt{woody} is \texttt{W UH D IY}. 
Each phoneme is also associated with one of the $8$ articulation manners shown in Table \ref{tab:manners}.

\begin{figure}[]
	\begin{subfigure}{0.475\columnwidth}
		\includegraphics[width=\columnwidth]{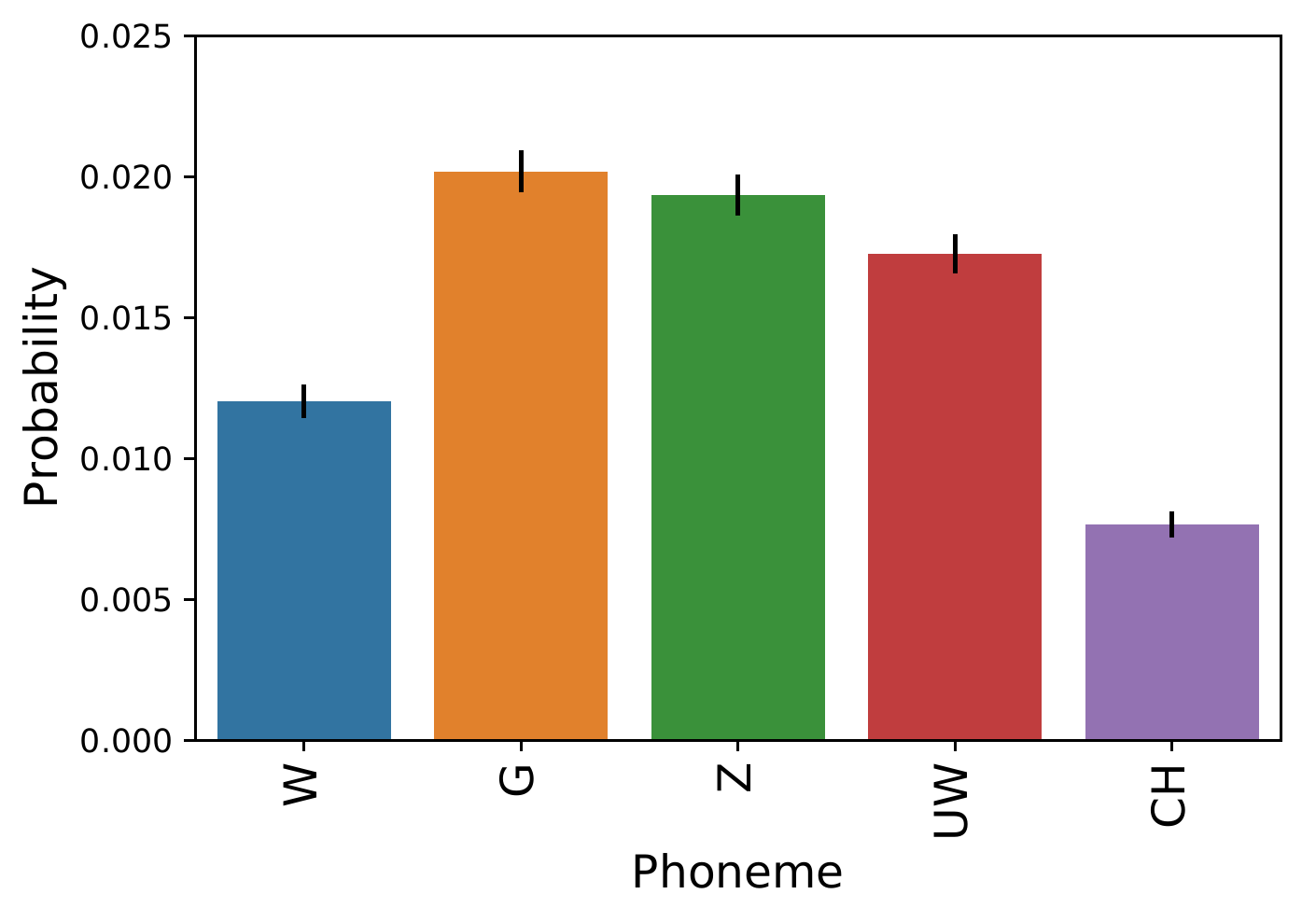}
		\caption{Slang}
		\label{fig:phonemes_slang_5}
	\end{subfigure}
	\begin{subfigure}{0.475\columnwidth}
		\includegraphics[width=\columnwidth]{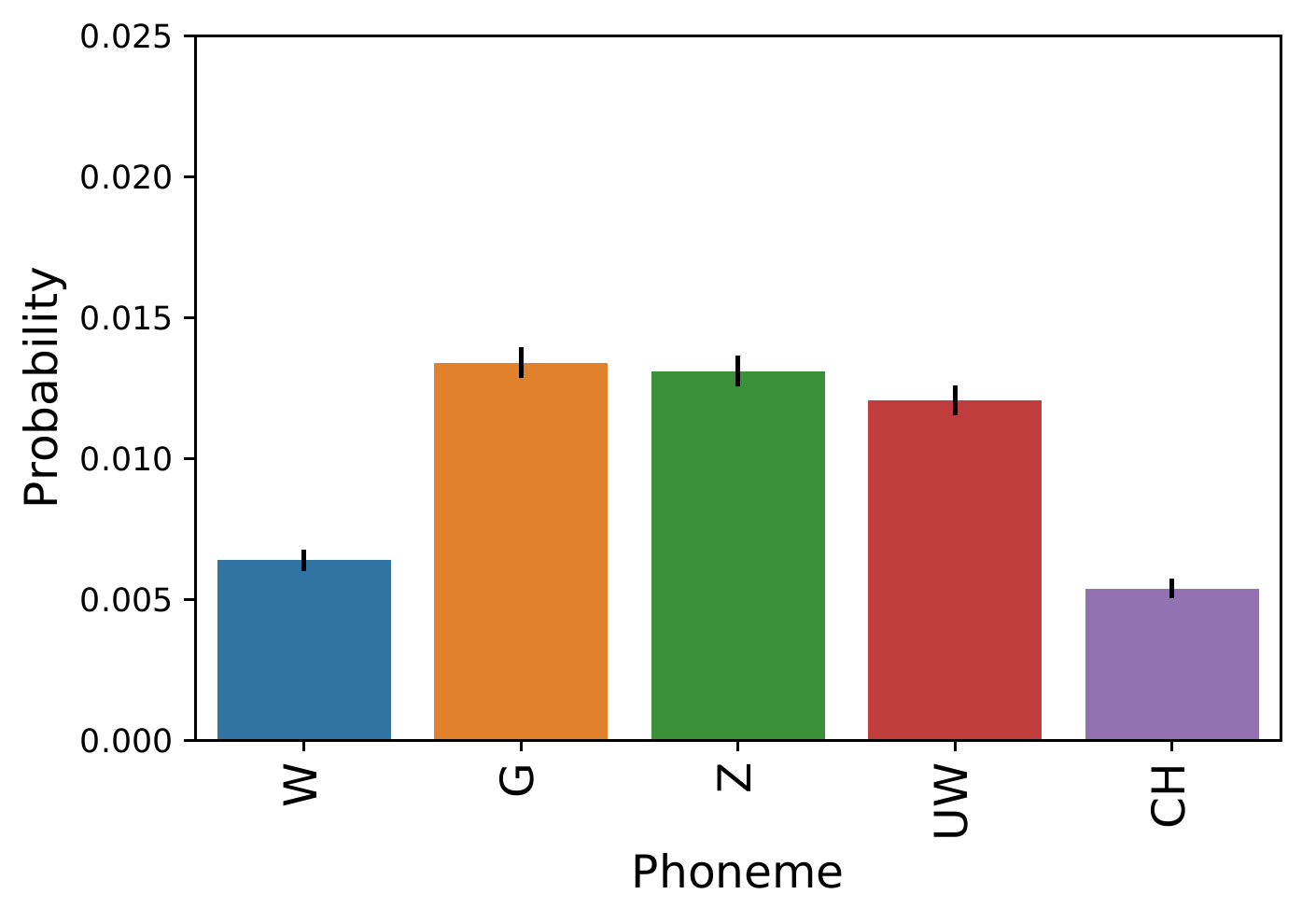}
		\caption{Standard English.}
		\label{fig:phonemes_standard_5}
	\end{subfigure}
	\caption{5 most over-represented phonemes in slang. Note the over-representation of \texttt{W, G, Z, UW and CH} in slang. We explain this by noting several slang words use phonemes like \texttt{Z} in words like \texttt{zucker, jizz} etc.}
	\label{fig:phonemes_5}
\end{figure}

\begin{table}[]
\centering
	\begin{tabular}{l|p{6cm}}
		\hline
		\textbf{Manner} & \textbf{Phonemes}  \\
		\hline
		\textsc{Stop} & B, D, G, K, P, T \\
		\textsc{Fricative} & DH, F, S, SH, TH, V, Z, ZH \\
		\textsc{Vowel} & AA, AE, AH, AO, AW, AY, EH, ER, EY, IH, IY,OW, OY, UH, UW  \\
		\textsc{Nasal} & M, N, NG \\
		\textsc{Liquid} & L, R \\
		\textsc{Affricate} & CH, JH \\
		\textsc{Aspirate} & HH \\
		\textsc{Semivowel} & W, Y \\
		\hline
	\end{tabular}
	\caption{Manners of Articulation for phonemes as per \emph{CMUDict}.}
	\label{tab:manners}
\end{table}

\textbf{Are certain phonemes over-represented in slang?}
To answer this question, we estimate the distribution over phonemes for words in our data-sets and rank the phonemes in descending order of their odds ratio (between slang and Standard English).  Figure \ref{fig:phonemes_5} shows the top $5$ phonemes in slang that are over-represented. 
In particular, note the presence of phonemes like \texttt{W, G and Z} which are used less frequently in Standard English. Examples of slang that uses these phonemes are \texttt{zazzed, zucker, pizzle, fonzie, woodie, faggoth} suggesting evidence of phonological variation between slang and Standard English.
 
\begin{figure}[]
	\begin{subfigure}{0.475\columnwidth}
		\includegraphics[width=\columnwidth]{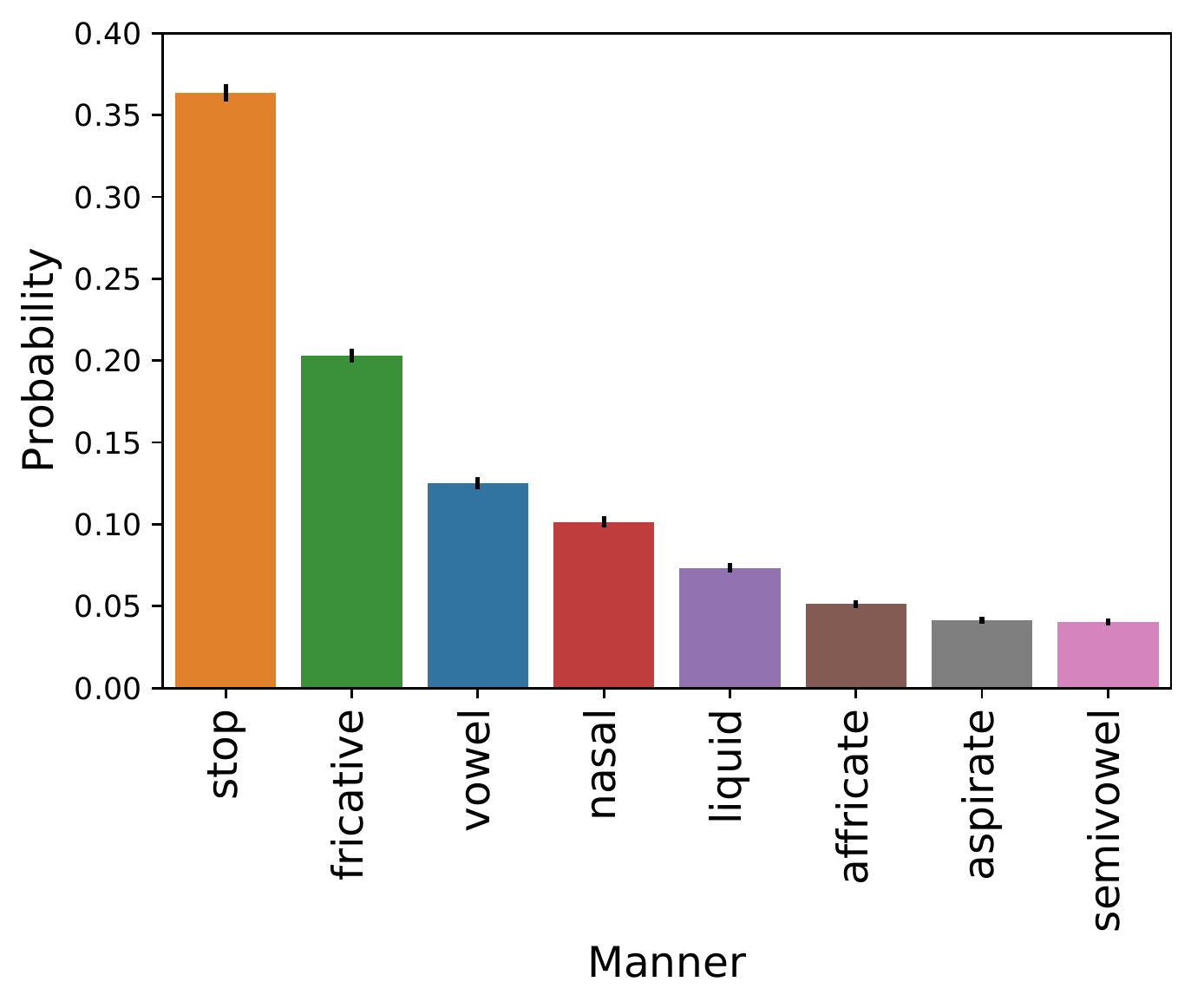}
		\caption{First Phoneme in slang}
		\label{fig:phonemes_prefixes_slang}
	\end{subfigure}
	\begin{subfigure}{0.475\columnwidth}
		\includegraphics[width=\columnwidth]{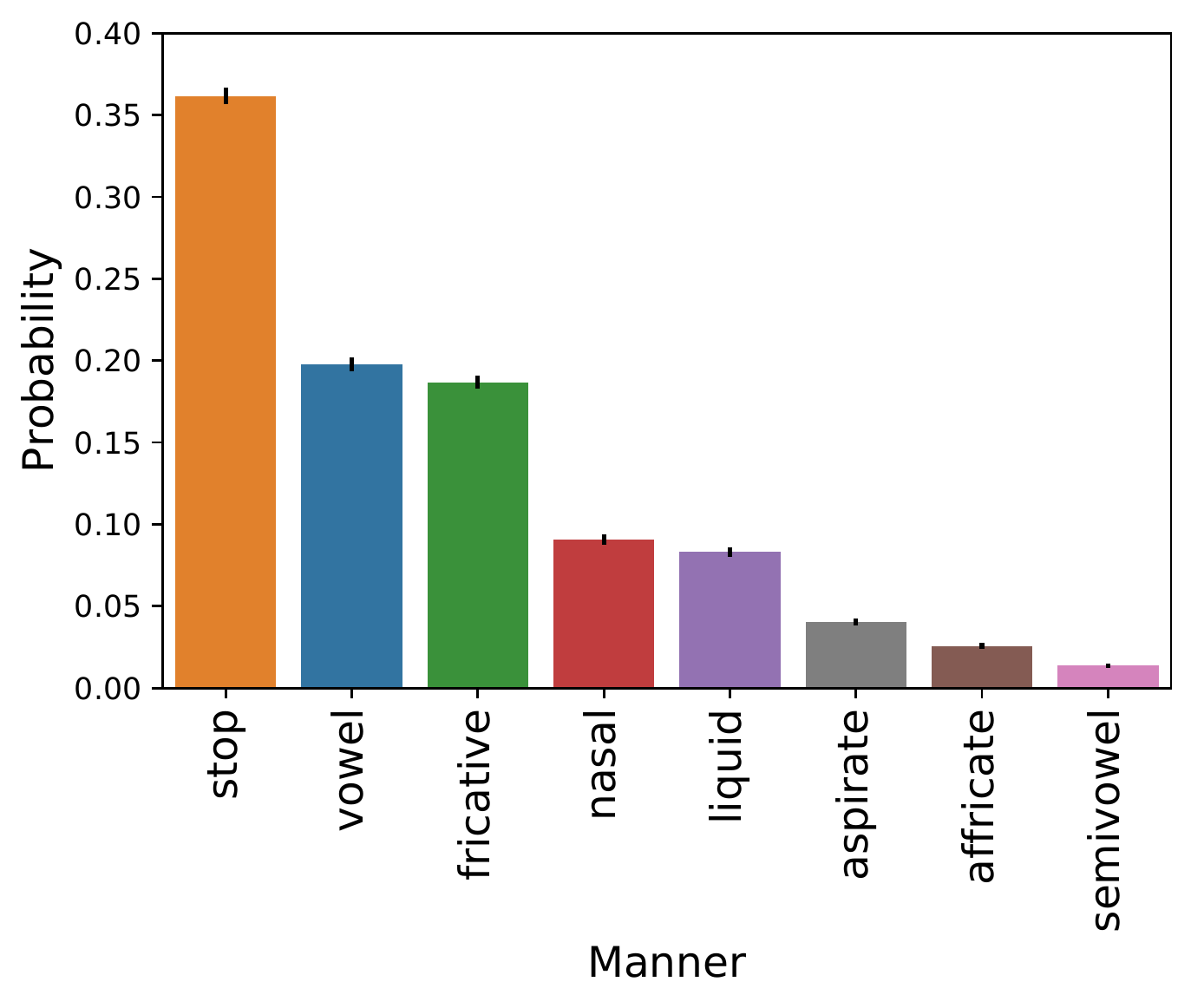}
		\caption{First Phoneme in SE}
		\label{fig:phonemes_prefixes_std}
	\end{subfigure}	
	\begin{subfigure}{0.475\columnwidth}
		\includegraphics[width=\columnwidth]{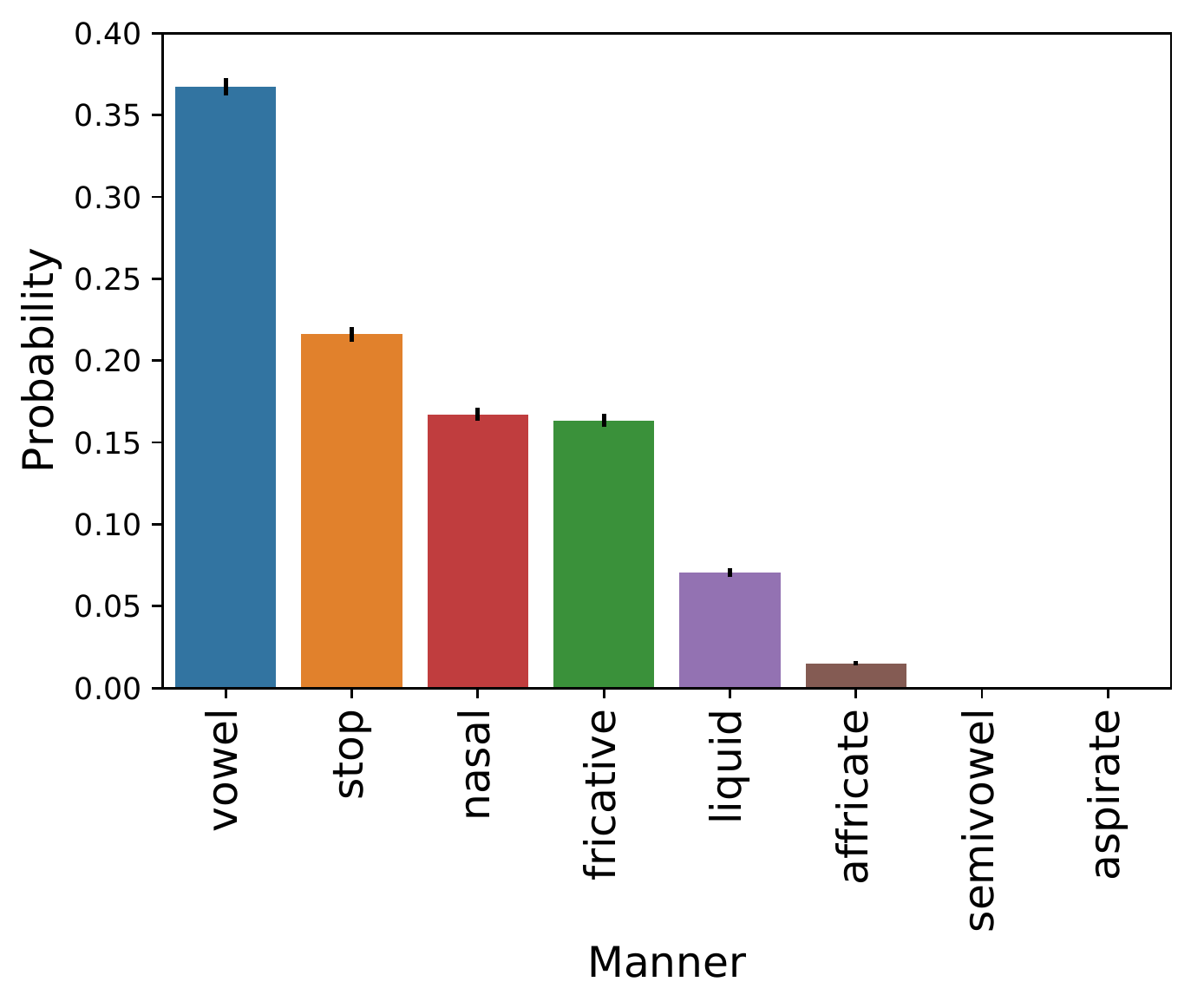}
		\caption{Final Phoneme in slang}
		\label{fig:phonemes_suffixes_slang}
	\end{subfigure}
	\begin{subfigure}{0.475\columnwidth}
		\includegraphics[width=\columnwidth]{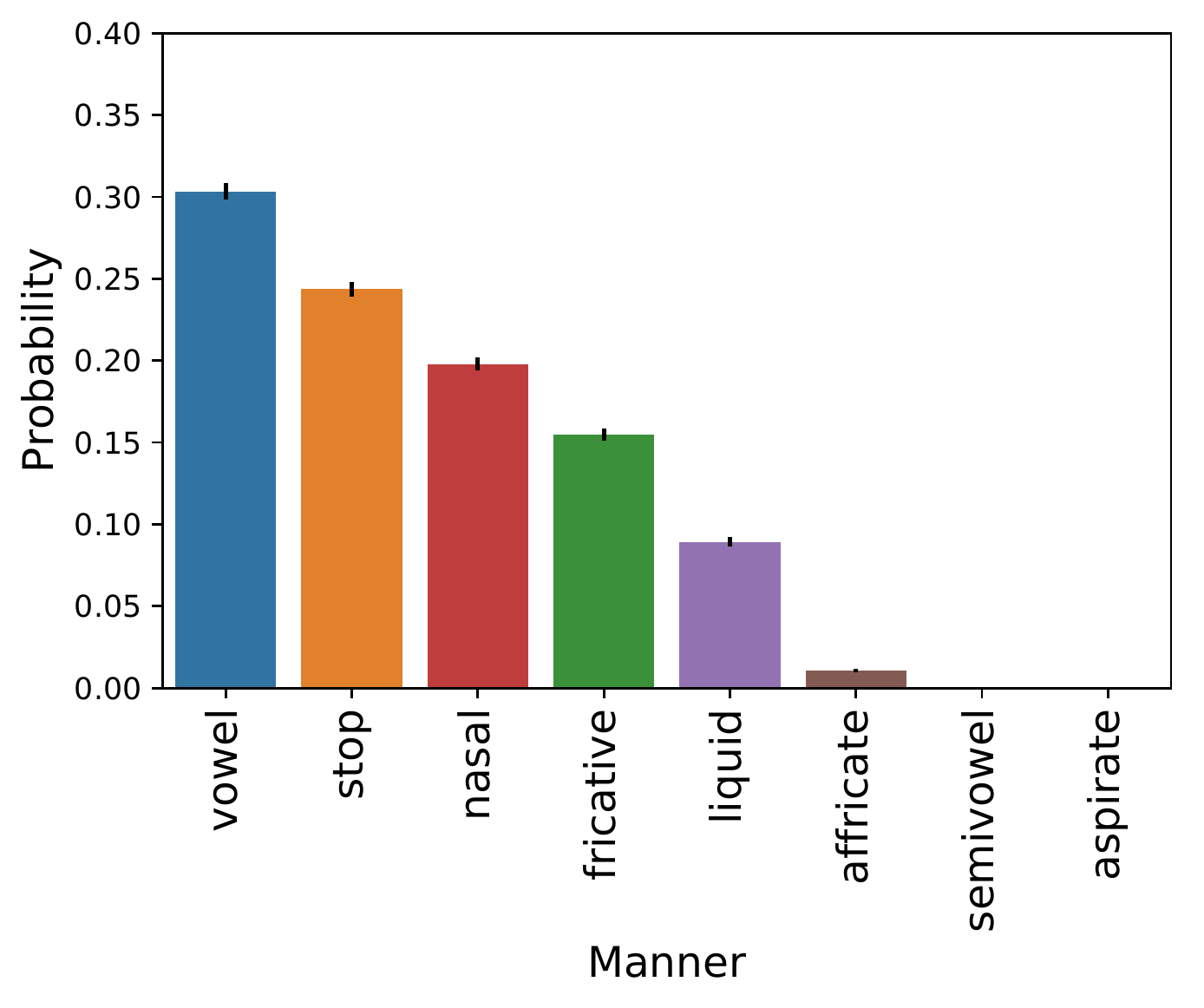}
		\caption{Final Phoneme in SE}
		\label{fig:phonemes_suffixes_std}
	\end{subfigure}
	\caption{Articulation Manners in slang and Standard English(SE). Observe significant differences for the first and final phoneme in slang when contrasted with the standard form. For example, \textsc{Fricatives} are more common than \textsc{Vowels} as the first phoneme in slang. Similarly \textsc{Affricates} as the first phoneme are more common in slang than in Standard English (for example: \texttt{chemtard, chadzing}).} 
	\label{fig:phonemes}
\end{figure}

\textbf{Does slang differ in manners of articulation?}
Figure \ref{fig:phonemes} shows the distribution of articulation manners obtained for the first and final phonemes in both slang and Standard English (SE).
First, consider the distribution in the first phoneme (Figures \ref{fig:phonemes_prefixes_slang} and \ref{fig:phonemes_prefixes_std}). 
We observe the following: (a) In slang, \textsc{Fricatives} as the first phoneme are more common than \textsc{Vowels}. We explain this by noting that several slang words begin their pronunciation using the fricative phonemes: \texttt{S} as in \textit{selfy}, \textsc{F} as in \textit{flub} and \textsc{SH} as in \textit{schmammered}. (b) Second, the proportion of words in slang whose first phoneme is a vowel is lower by $7.2\%$ points than Standard English (SE). (c) Finally, words which begin with \textsc{Affricates}: \textsc{CH, JH} are  much more common in slang than in Standard English. Examples of such slang words are \texttt{chemtard, chelly, chicklet, charva, chadzing, juvy, jerkweed}. Additionally, we noted no statistically significant difference in the proportions of \textsc{Stop} phoneme as the first phoneme between slang and Standard English.\footnote{Highlighted significant differences in proportions were statistically significant at $\alpha=0.05$ using z-test for proportion differences with Bonferroni correction.} We now turn our attention to manner distribution of the final phoneme (Figures \ref{fig:phonemes_suffixes_slang} and \ref{fig:phonemes_suffixes_std}). 
Observe that the final phoneme being a \textsc{Vowel} is more common (by $6.3\%$ points) in slang than standard English and also note a decrease in \textsc{Stop} phonemes by $\sim2.7\%$ points in slang. 

\textbf{Conclusion} To summarize, our analysis reveals that phonological properties of slang are markedly different from words in standard English where certain phonemes and manners of articulation like \textsc{Fricatives} and \textsc{Affricates} are over-represented in slang.

\subsection{Morphology}
While morphological patterns and derivations are widely studied for standard forms of English \cite{aronoff1976word,bauer1983english,scalise1986generative,mayerthaler1981morphologische},  \citet{mattiello2008introduction} observes that many formations of slang have been largely ignored since they are far displaced from the regular word formation patterns and thus \emph{extra-grammatical}, with only a small focus on standard word formation rules in slang \cite{eble2012slang}. Consequently, \cite{mattiello2008introduction} claims that studying the expressive morphological characteristics of slang can shed light not only on the creative process of language formation, grammar formation but also provide insight into its semantics and sociological impact.

In line with this viewpoint, we now analyze the morphological patterns of slang. 
Our analysis yields insight into the morphological patterns evident in formation of different classes of slang like blends, clippings and reduplicatives. 

\subsubsection{\textbf{Analysis of Morphological Patterns}}

\textbf{How does slang differ in its morphological forms from words in standard dialect?}
To answer this, we decompose words into their corresponding morphemes, by learning a model to segment words into morphemes using \textsc{Morfessor} \cite{virpioja2013morfessor}\footnote{We use the default set of parameters for \textsc{Morfessor}.}.

\begin{figure}[]
	\begin{subfigure}{0.475\columnwidth}
		\includegraphics[width=\columnwidth]{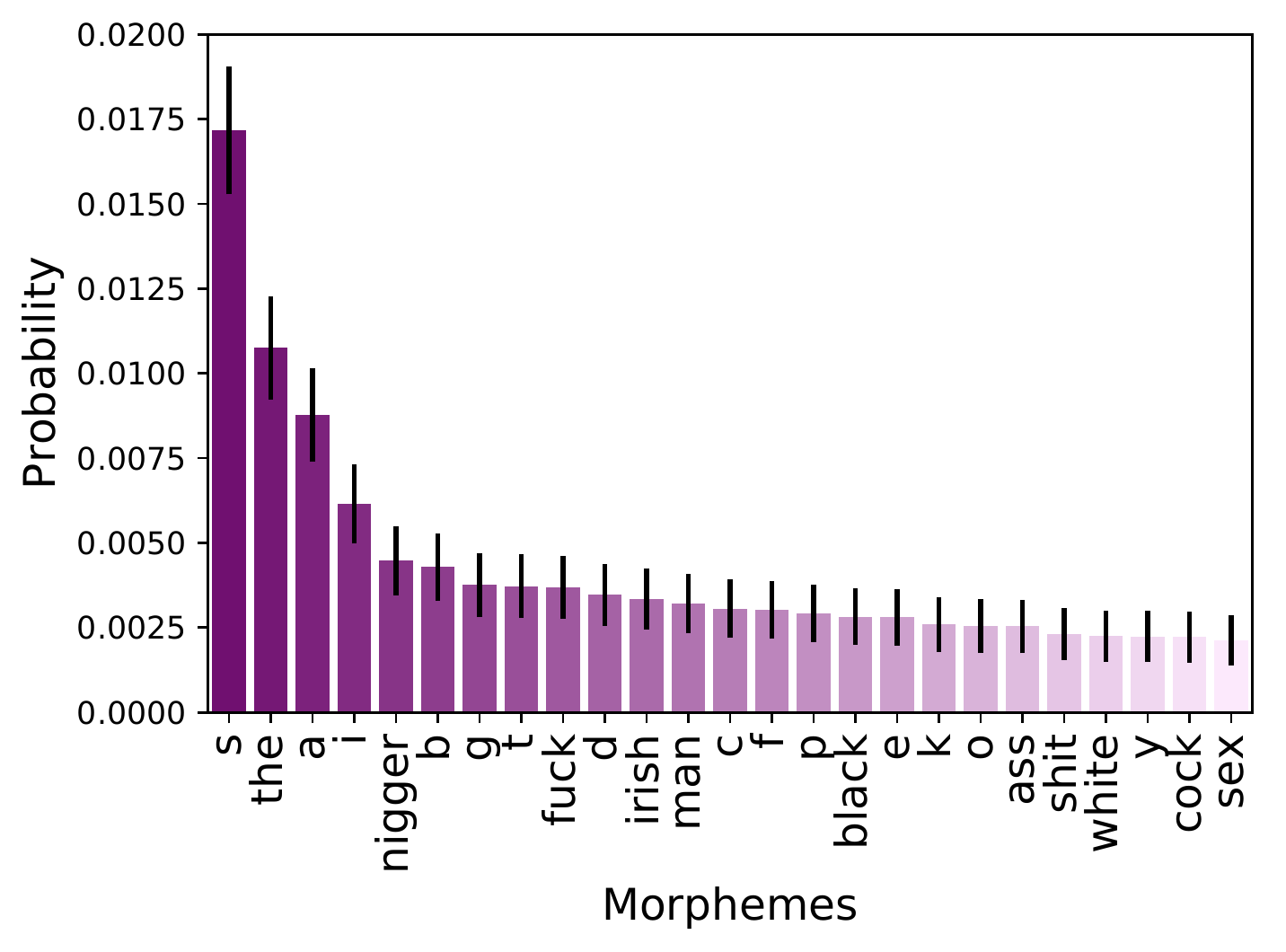}
		\caption{Top $25$ prefixes in slang}
		\label{fig:prefixes_slang_25}
	\end{subfigure}
	\begin{subfigure}{0.475\columnwidth}
		\includegraphics[width=\columnwidth]{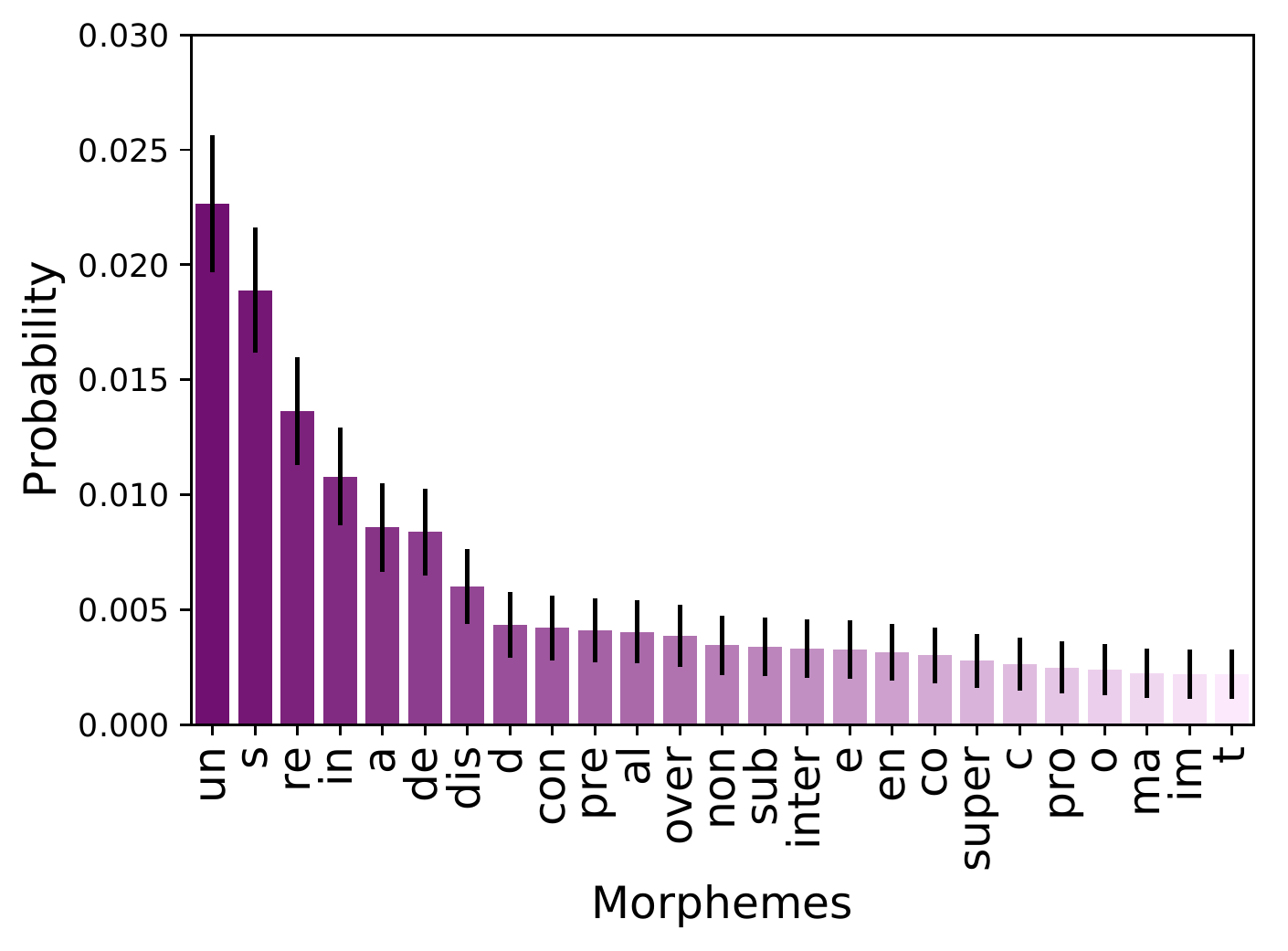}
		\caption{Top $25$ prefixes in SE.}
		\label{fig:prefixes_standard_25}
	\end{subfigure}
\begin{subfigure}{0.475\columnwidth}
	\includegraphics[width=\columnwidth]{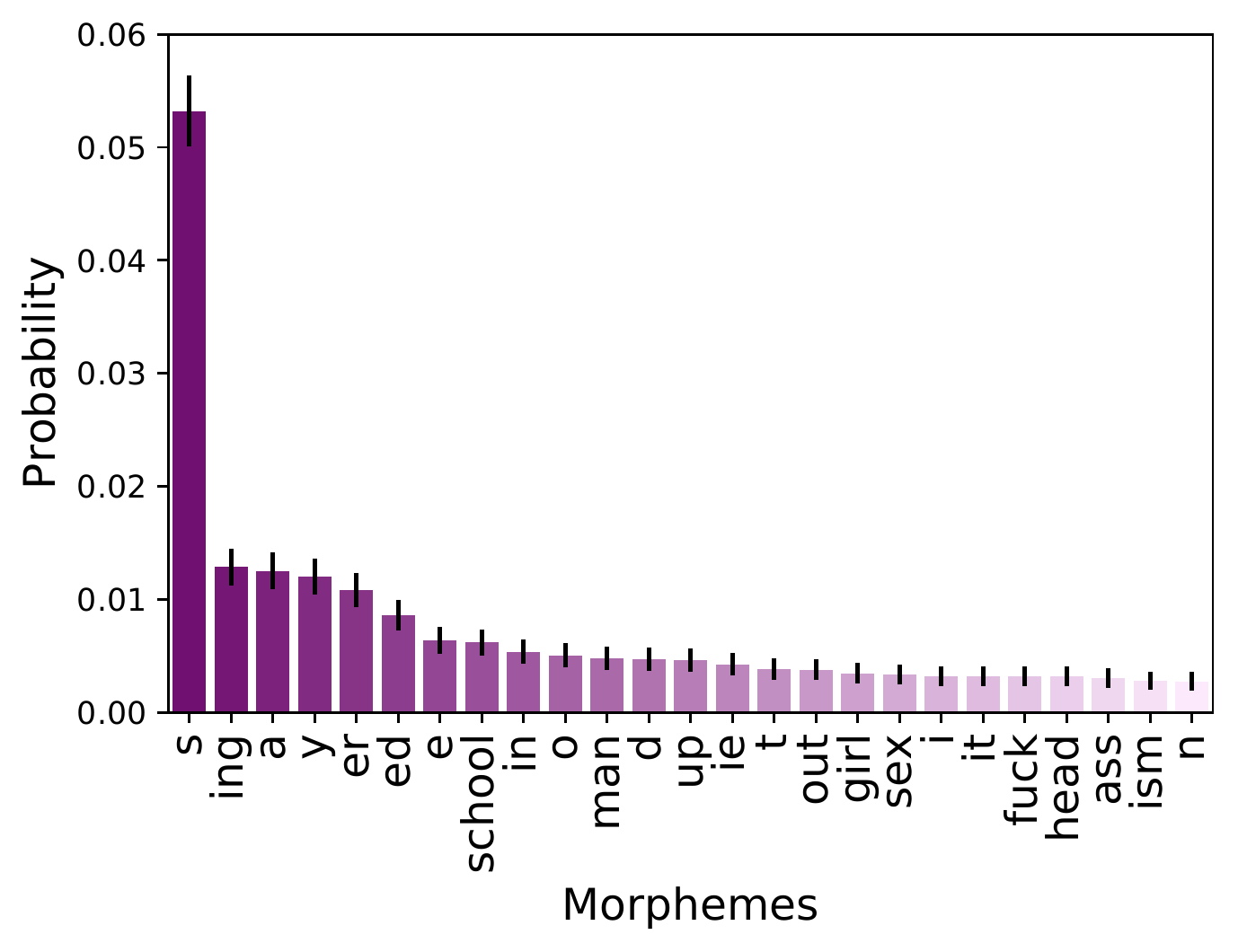}
	\caption{Top $25$ suffixes in slang}
	\label{fig:suffixes_slang_25}
\end{subfigure}
\begin{subfigure}{0.475\columnwidth}
	\includegraphics[width=\columnwidth]{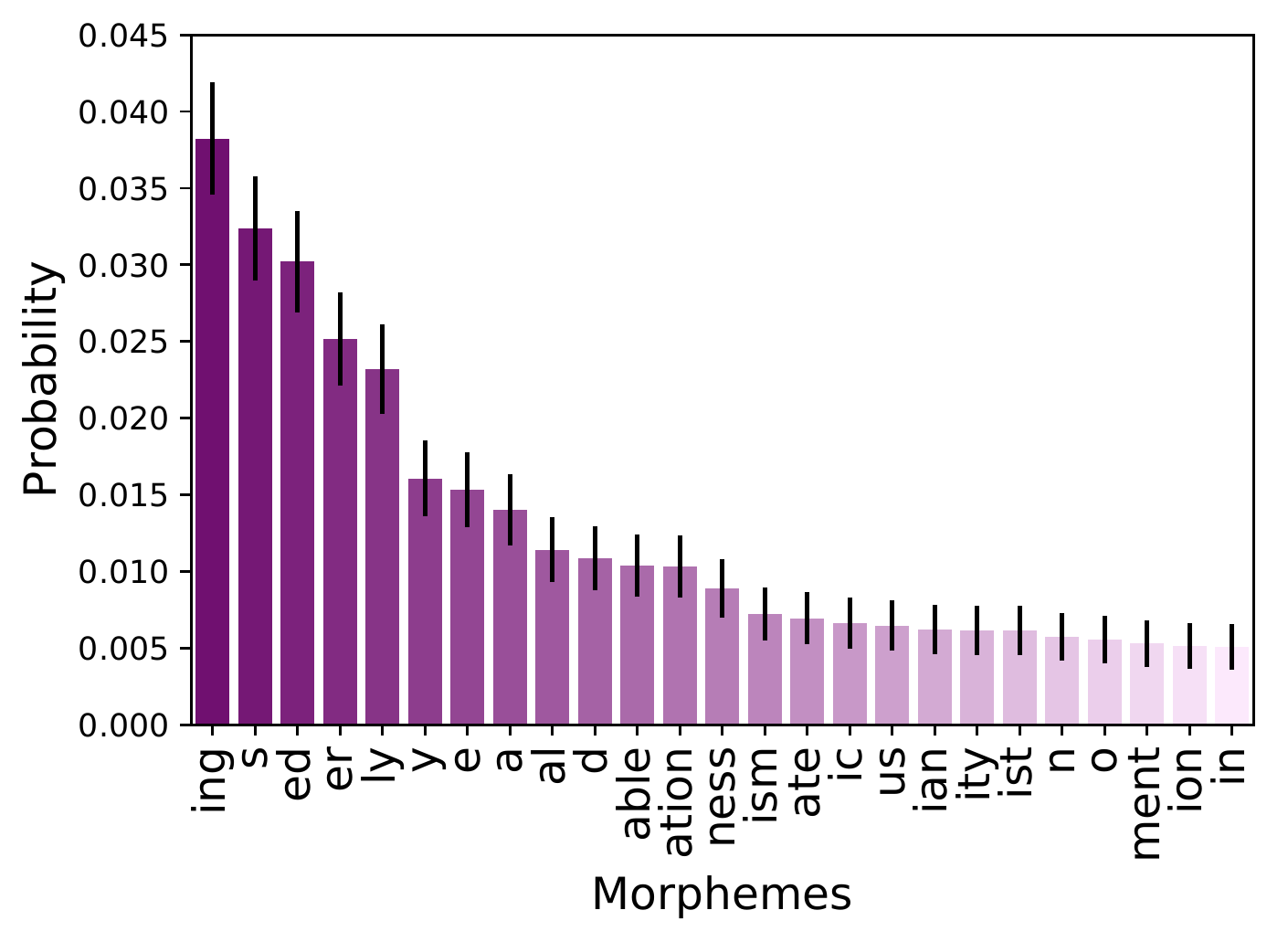}
	\caption{Top $25$ suffixes in SE.}
	\label{fig:suffixes_standard_25}
\end{subfigure}
\caption{Morphological differences in slang and Standard English (SE). Slang exhibits morphological derivations quite distinctive from Standard English, using non-standard suffixes and prefixes like \texttt{fuck, nigger, ass}.}
\label{fig:morph}
\end{figure}

Figure \ref{fig:morph} shows the top $25$ most common prefixes and suffixes in both slang and standard English. We immediately make these observations: First, slang demonstrates a heavier tail than Standard English. In slang, the top $25$ prefixes only account for $10\%$ of the total mass where as the top $25$ prefixes in Standard English account for $15\%$ suggesting a rich and more varied word formation in slang. Second, note the much higher mass assigned to  non-standard prefixes in slang like \texttt{nigger, fuck, irish, man, black, ass, shit, white, cock and sex}.
Similar observations are noted in the case of morphological suffixes as well by noting the presence of non-standard suffixes like \texttt{school, man, out, girl, sex, fuck, head and ass.}

\textbf{Conclusion}: Our large scale computational analysis is consistent with the observations made by \cite{mattiello2013extra} who notes that morphology in slang exhibits extra-grammatical rules which are not observed in word formation in standard English. Moreover, we quantitatively estimate the likelihoods of these patterns (both prefixes and suffixes) characterizing the heavy tail nature of such patterns in slang. 

\begin{figure}
\begin{subfigure}{0.45\columnwidth}
	\includegraphics[width=0.9\columnwidth]{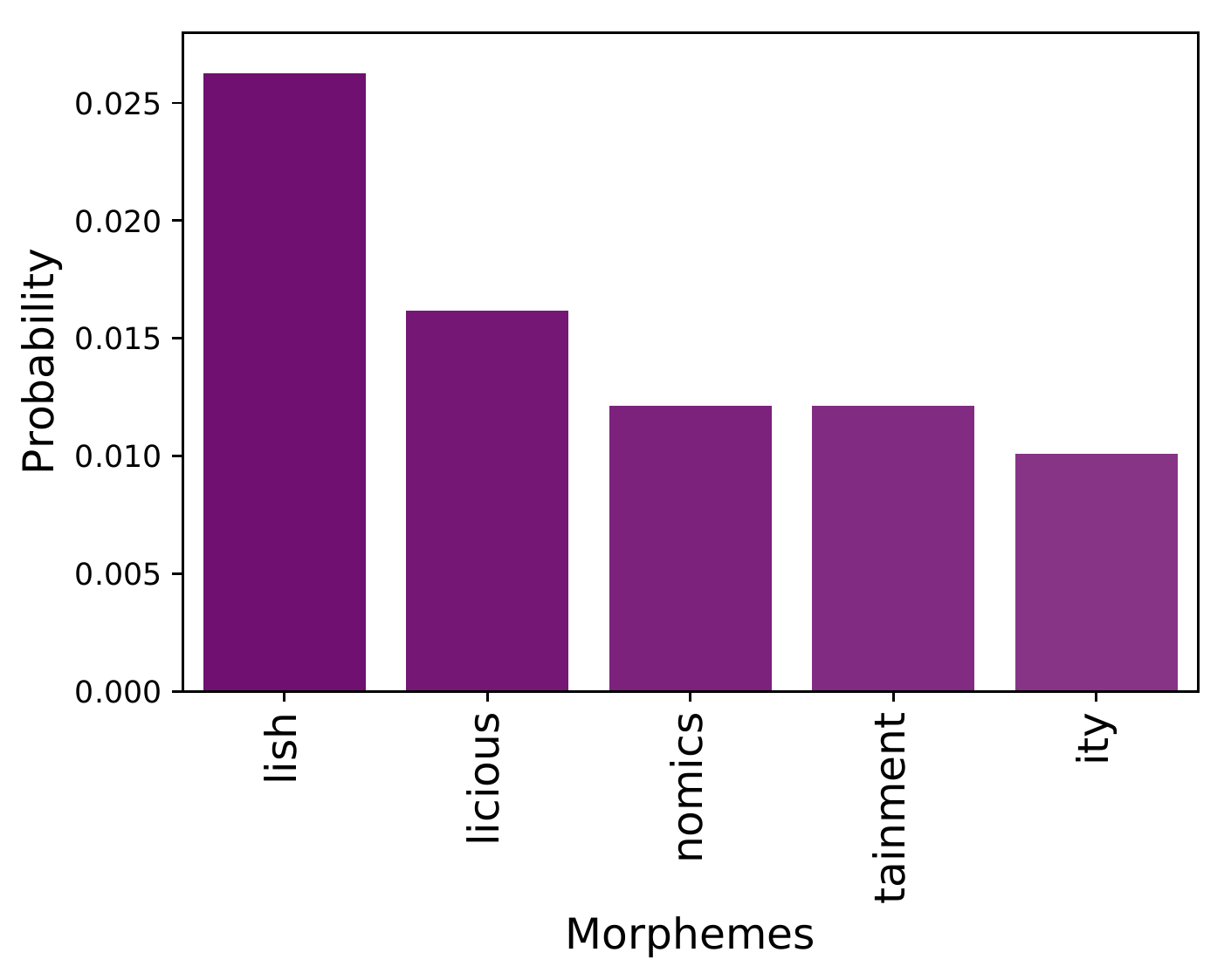}
	\captionsetup{width=.9\linewidth}
	\caption{Top 5 suffixes in slang blends. Note these are not dominant suffixes in SE.}
	\label{fig:blends_suffixes}
\end{subfigure}
\begin{subfigure}{0.45\columnwidth}
	\includegraphics[width=0.9\columnwidth]{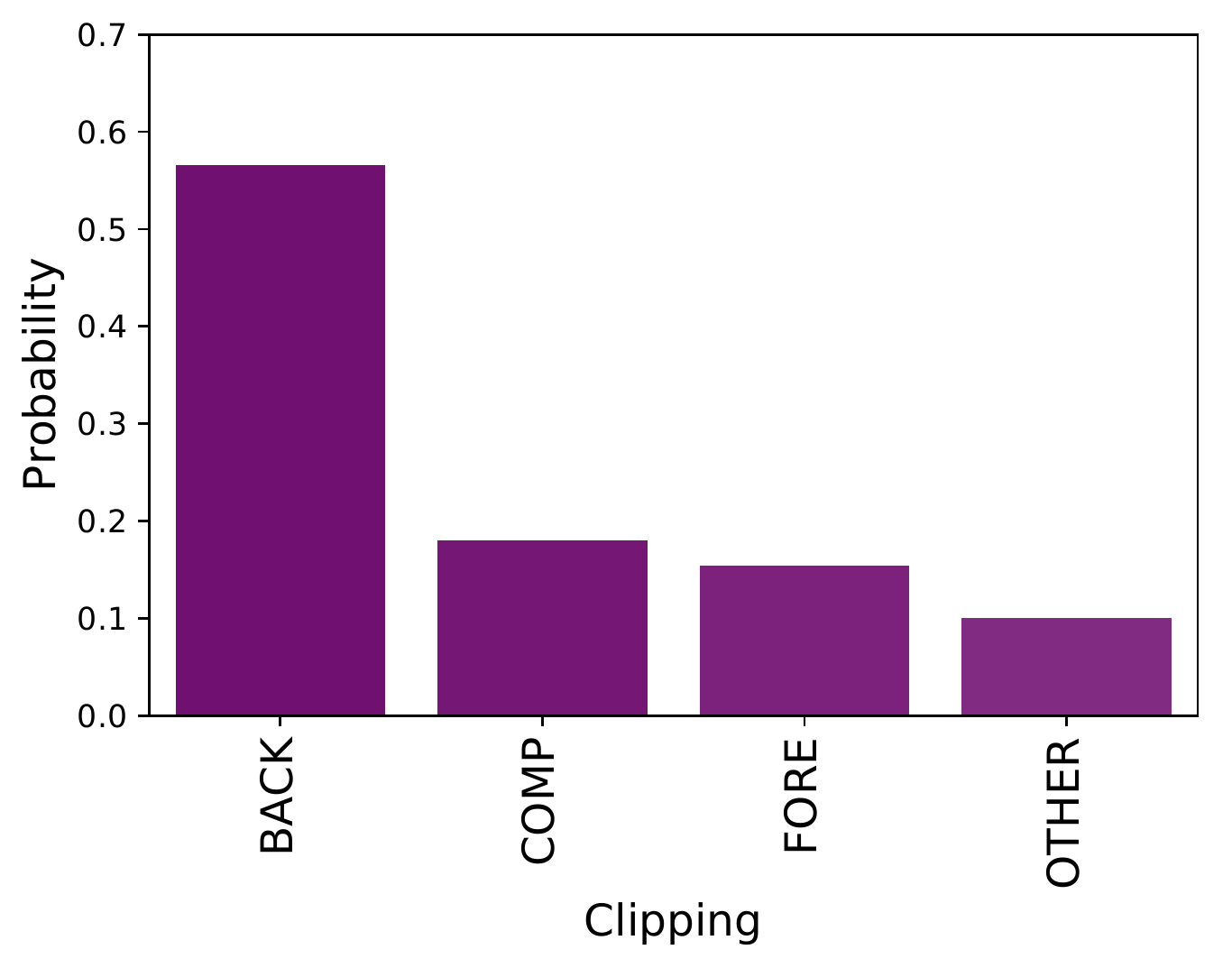}
	\caption{Most clippings are \textsc{Back} (nigg) clippings and \textsc{Compound} clippings (slowmo). }
	\label{fig:clippings_stype}
\end{subfigure}
\caption{Linguistic patterns of Blends and Clippings.}
\end{figure}

\begin{figure}[htb!]
	\begin{subfigure}{0.475\columnwidth}
		\includegraphics[width=\columnwidth]{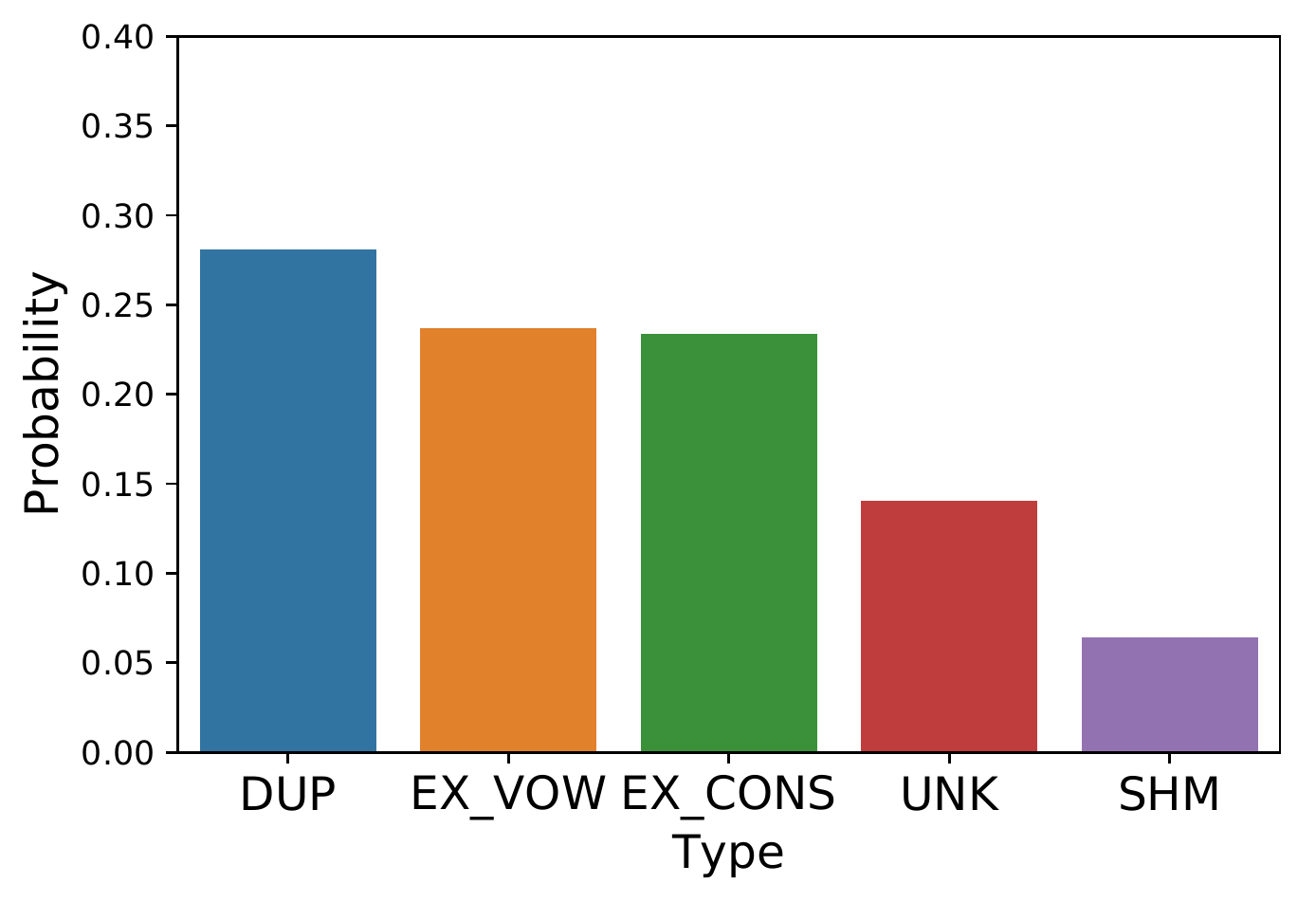}
		\caption{Top Reduplicative patterns. Duplicate is most common.}
		\label{fig:top_redups}
	\end{subfigure}
	\begin{subfigure}{0.475\columnwidth}
		\includegraphics[width=\columnwidth]{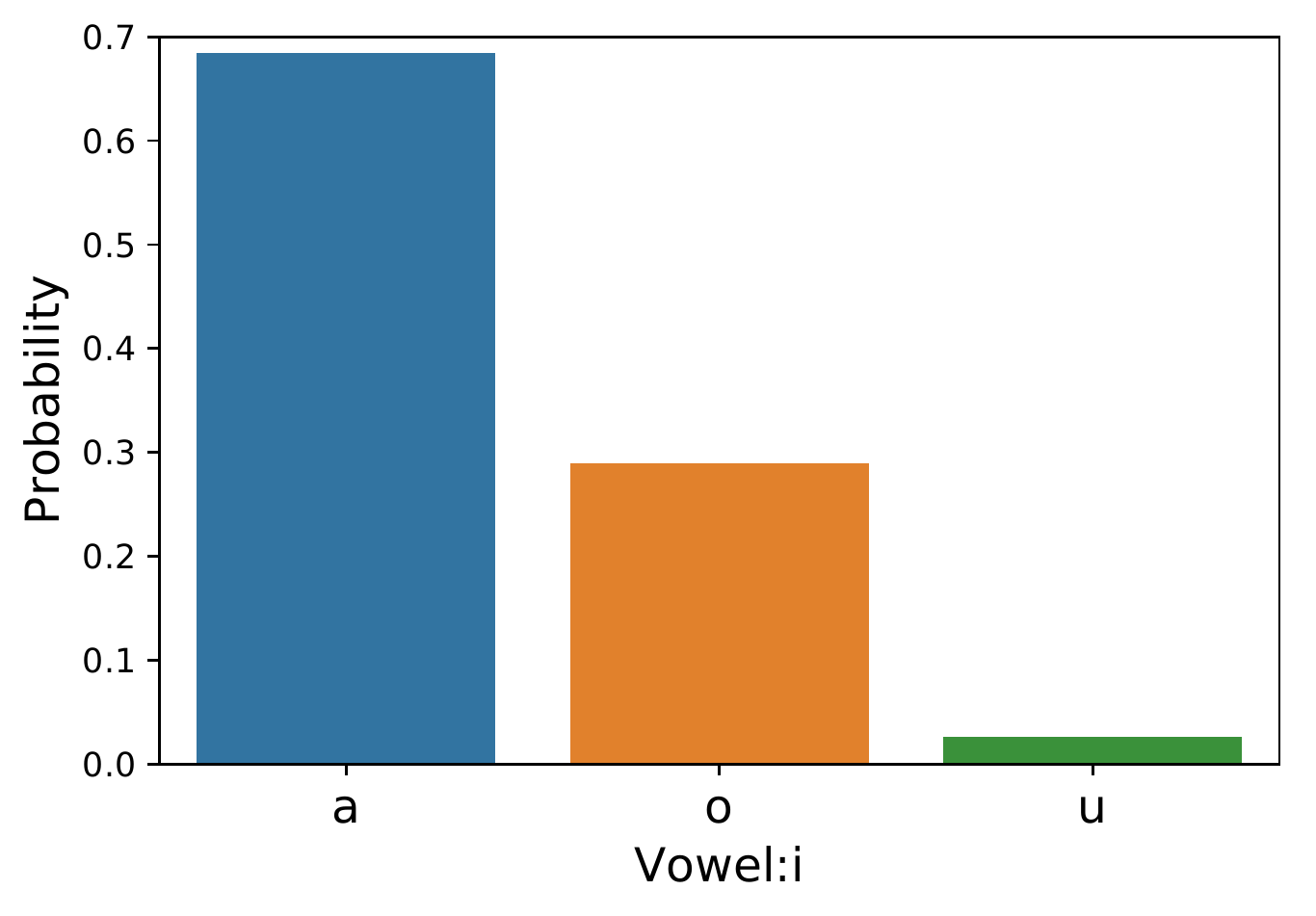}
		\caption{Top letters which replace 'i'. (bing-bang)}
		\label{fig:rep_i}
	\end{subfigure}
	\begin{subfigure}{0.475\columnwidth}
		\includegraphics[width=\columnwidth]{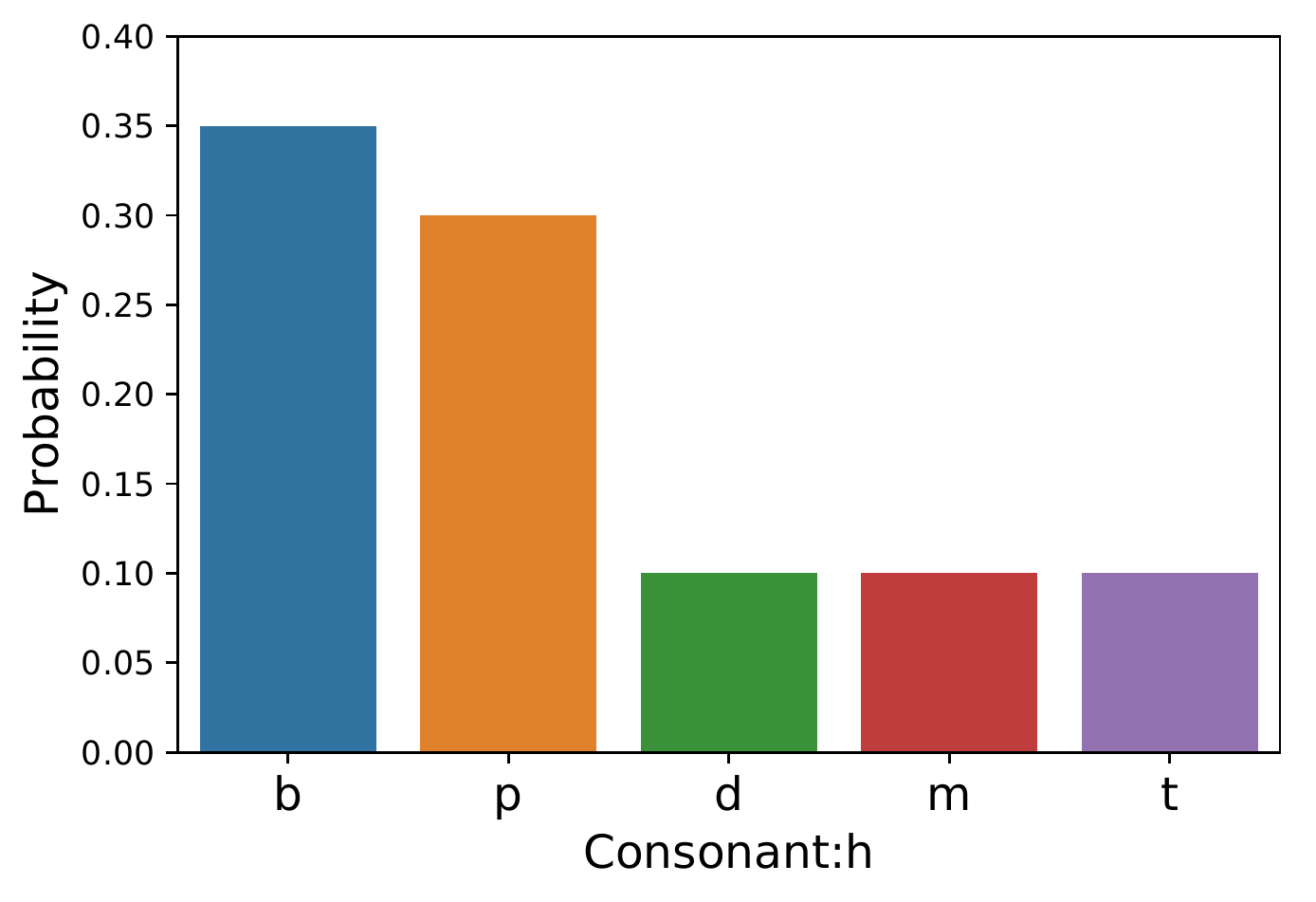}
		\captionsetup{width=.9\linewidth}
		\caption{Top letters which replace 'h'. (hurly-burly)}
		\label{fig:rep_h}
	\end{subfigure}
	\begin{subfigure}{0.475\columnwidth}
		\captionsetup{width=.9\linewidth}
		\includegraphics[width=\columnwidth]{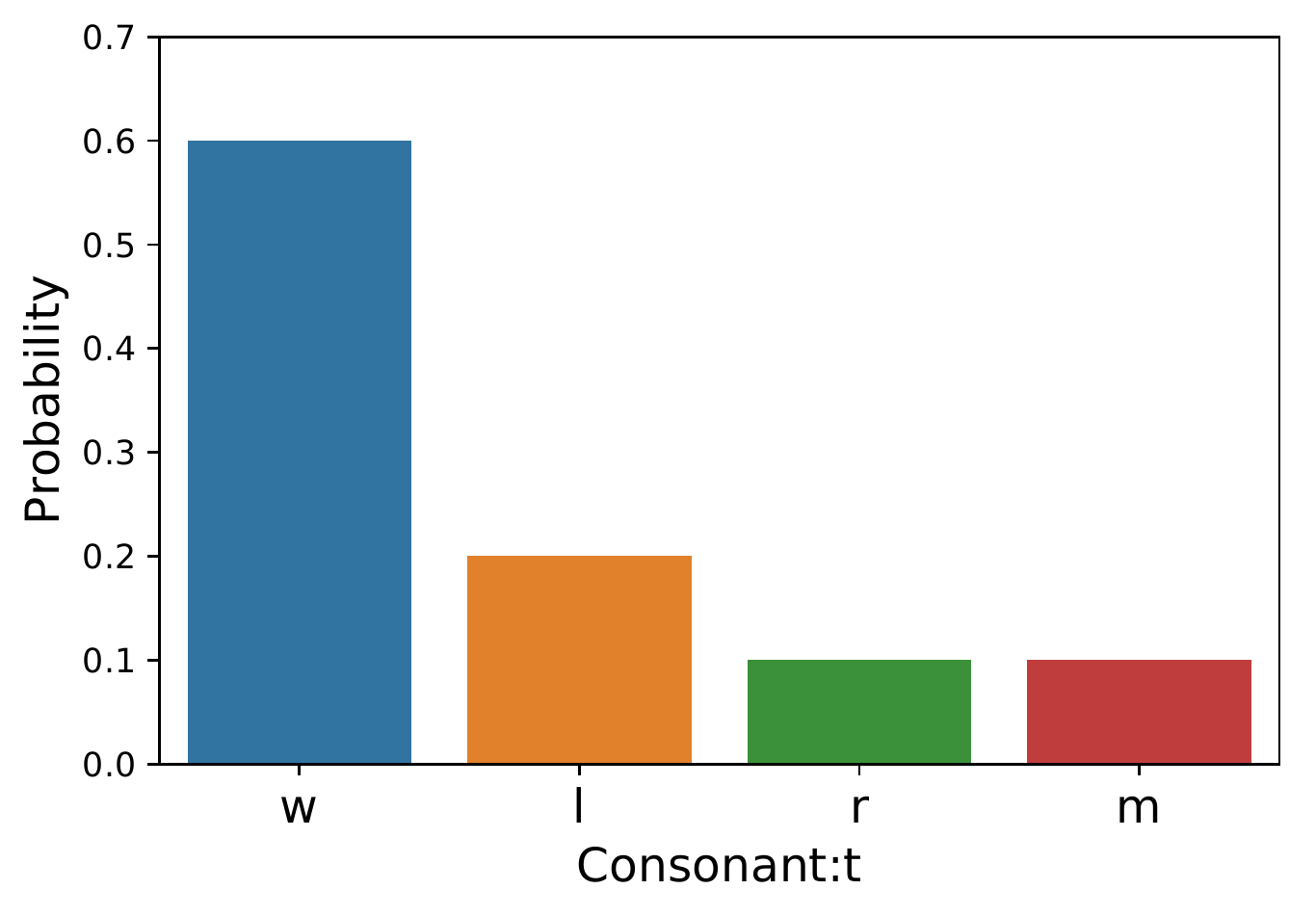}
		\caption{Top letters which replace 't'. (teenie-weenie).}
		\label{fig:rep_t}
	\end{subfigure}
	\caption{Patterns in reduplicative formation.}
	\label{fig:redups_patterns}
\end{figure}

\subsubsection{\textbf{Slang Classes}}
\label{sec:slang_classes}
Having analyzed morphological patterns of slang broadly, we now turn our attention to $4$ specific categories of slang identified by \citet{mattiello2013extra} that ``exhibit underlying preferences for some underlying morphological patterns''. We describe these classes briefly and then thoroughly analyze morphological patterns for each of these classes revealing insights into their generative process.
\begin{itemize}
	\item \textbf{Alphabetisms} are shortenings of a multi-word sequence (for eg. \texttt{hmu} from \texttt{hit me up} or \texttt{TV} from \texttt{tranvestite}). 
	Alphabetisms can be sub-categorized into two types based on their pronunciation although the distinction may not always be clear: (a) \emph{Acronyms} are pronounced by using the regular reading rules (for eg. dink) (b) \emph{Initialisms} are pronounced letter by letter (for eg. BLT). 
	While it may appear that construction of alphabetisms is very predictable, it is not necessarily the case. For example, as \citet{mattiello2013extra} notes, \texttt{University of the Arts London} could be abbreviated as \texttt{UAL} or \texttt{UOTAL}.
	\item \textbf{Blends} are formed by merging parts of existing words. For example: \texttt{sextini} is formed by merging parts of \texttt{sex} and \texttt{martini}. \citet{mattiello2013extra} observes that even though blends are ubiquitous in slang, they do not exhibit strict rules of formation but instead only show affinities for some patterns which we characterize quantitatively. 
	\item \textbf{Clippings} are obtained by shortening a lexeme into a small number of syllables. For eg. \texttt{fave} is a clipping of \texttt{favorite} and \texttt{gym} is a clipping of \texttt{gymnasium}.
	Clippings can be further classified into 3 major classes depending on the portion that is being clipped: (a) \textsc{Back} clipping where the beginning of the word (lexeme) is retained (like \texttt{nigg} from \texttt{nigger}) (b) a \textsc{Fore} clipping, where the end of a word is retained (like \texttt{roach} from \texttt{cockroach}) and (c) A compound clipping (\texttt{slowmo}) is a clipping of a compound word (\texttt{slow motion}). 
	\item \textbf{Reduplicatives} (also called echo-words or flip-flop words) are word pairs obtained by either repeating a word (\texttt{boo boo}) or by alternating certain vowels or consonants so that they are phonologically similar (\texttt{teenie weenie}). 
\end{itemize}
Finally while \cite{mattiello2013extra} provides a manually compiled list of $1580$ words belonging to each of the $4$ different classes mentioned, it is important to note that these categories are not exhaustive and several slang words do not fall into any of these categories (for eg. \texttt{edging}). 

\paragraph{{Morphological patterns of Slang Classes}}
 Here, we analyze common word patterns in (a) Blends (b) Clippings and (c) Reduplicatives, using the above gold standard dataset of $1580$ words.
\begin{itemize}
\item \textbf{Blends} Figure \ref{fig:blends_suffixes} shows the top 5 suffixes in blends compiled by \cite{mattiello2013extra,deri2015make}. Note the dominant presence of suffixes like \texttt{lish, licious, tainment} which yield a large number of blends like \texttt{Hinglish, bootylicious, fergilicious, sextainment, and infotainment} supporting \cite{mattiello2013extra} quantitatively that blends have affinities to certain suffixes. 
\item \textbf{Clippings} Figure \ref{fig:clippings_stype} shows the fraction of each clipping type. Most clippings are \textsc{Back} clippings while \textsc{Fore} clippings are relatively rare. 
\item \textbf{Reduplicatives} Figure \ref{fig:redups_patterns} shows preferred patterns of reduplicative formation.
First, observe that reduplicatives are pre-dominantly formed by one of the following processes: (a) \emph{Duplication (DUP)} like \texttt{boo-boo} (b) \emph{Exchanging a vowel (EX\_VOW)} like \texttt{flip-flop} and (c) \emph{Exchanging a consonant (EX\_CONS)} like \texttt{bitsy-witsy} while a small fraction of reduplicatives are formed by (d) \emph{Prefixing schm, shm (SHM)} like moodle-schmoodle and (e) other patterns (UNK). 
Second, vowel and consonant substitutions reveal dominant preferences. \texttt{i} is more likely to be substituted with \texttt{a} and \texttt{o} among other letters (Figure \ref{fig:rep_i}). \texttt{h} is much more likely to be substituted by \texttt{b} and \texttt{p} whereas \texttt{t} is much more likely to be substituted by \texttt{w} (Figures \ref{fig:rep_h} and \ref{fig:rep_t}). Examples are \texttt{bling-blang, shick-shack, flip-flop, hurly-burly, teenie-weenie}.
\end{itemize} 

\textbf{Conclusion} In summary, our analysis reveals varied patterns of formation of blends, clippings and reduplicatives.

\subsubsection{\textbf{Detection of Slang Classes}}
Having obtained insights into linguistic patterns governing four different slang categories, we demonstrate the utility of these insights in developing a predictive model to classify slang into one of these $4$ proposed categories. 
We evaluate our model quantitatively on a small gold-standard test set as well and apply our learned model to infer labels for a large list of words from \textsc{UrbanDictionary} to construct a much larger data-set to aid future qualitative and quantitative analysis. 

\paragraph{Gold standard Dataset} We once again use the data-set of $1580$ words by \cite{mattiello2013extra} as the gold standard data set for learning a classification model to classify a word into one of these four categories. We create a gold standard training and test data set by a random split using $10\%$ of the data as the test set.

\paragraph{Learning the model} We consider a simple Logistic Regression classifier and experiment with two sets of features:\footnote{Other models like SVM's also yielded similar performance.}
\begin{itemize}
	\item \textbf{Character N-gram features} of (length:1-5) where feature size is restricted to $200$.
	\item \textbf{Morphemes}: We consider morpheme-grams (1-5) as features and restrict our feature size to $200$. Given, the small amount of training data, we expect the morpheme feature set to be much sparser than the character ngram features and therefore expect a model learned using these features to perform worse than using character ngrams.
\end{itemize}
 Finally, as a baseline we consider a \emph{random} model which randomly draws predictions from the training data label distribution.

\begin{figure}
	\begin{subfigure}[t!]{0.475\columnwidth}
		\includegraphics[width=\columnwidth]{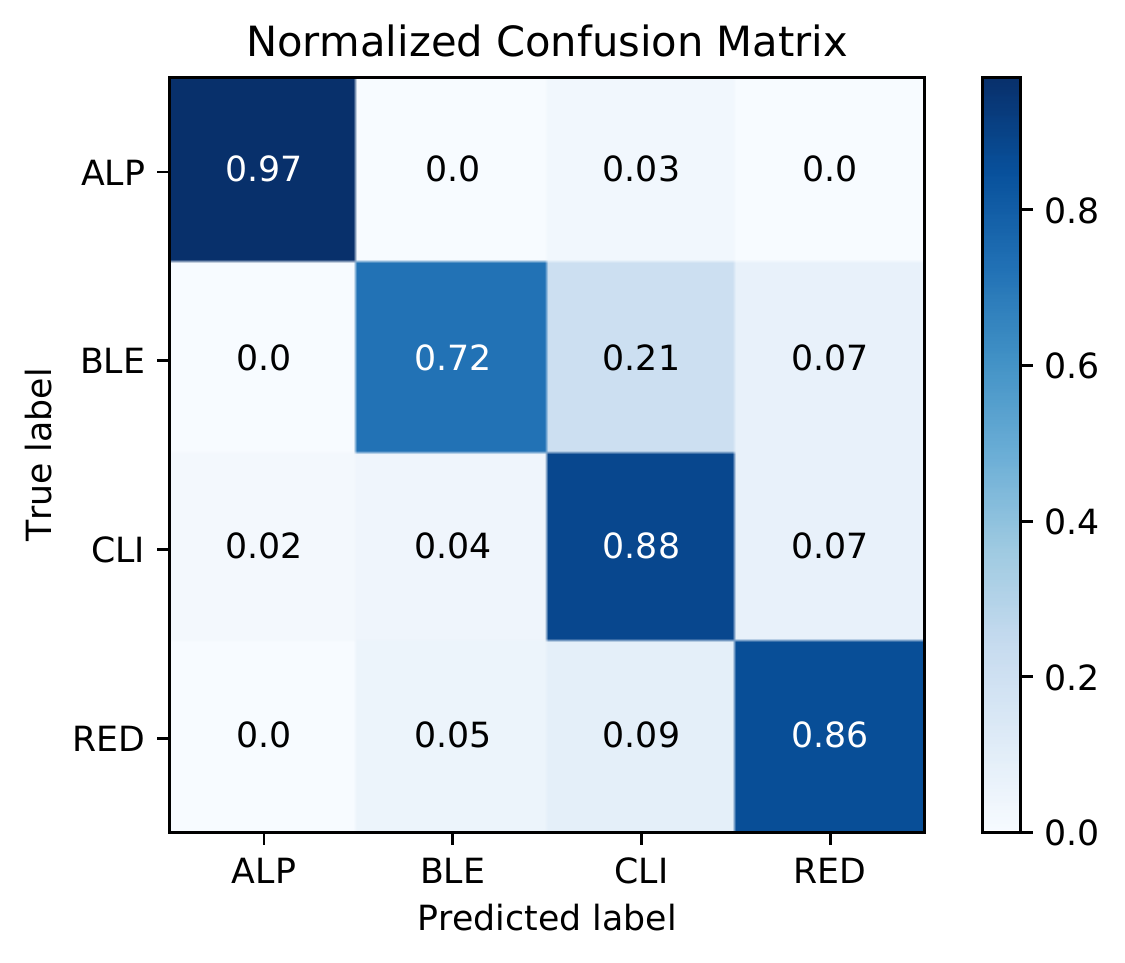}
		\caption{Confusion Matrix - Gold}
		\label{fig:morph_gold_standard}
	\end{subfigure}
	\begin{subfigure}[t!]{0.475\columnwidth}
		\includegraphics[width=\columnwidth]{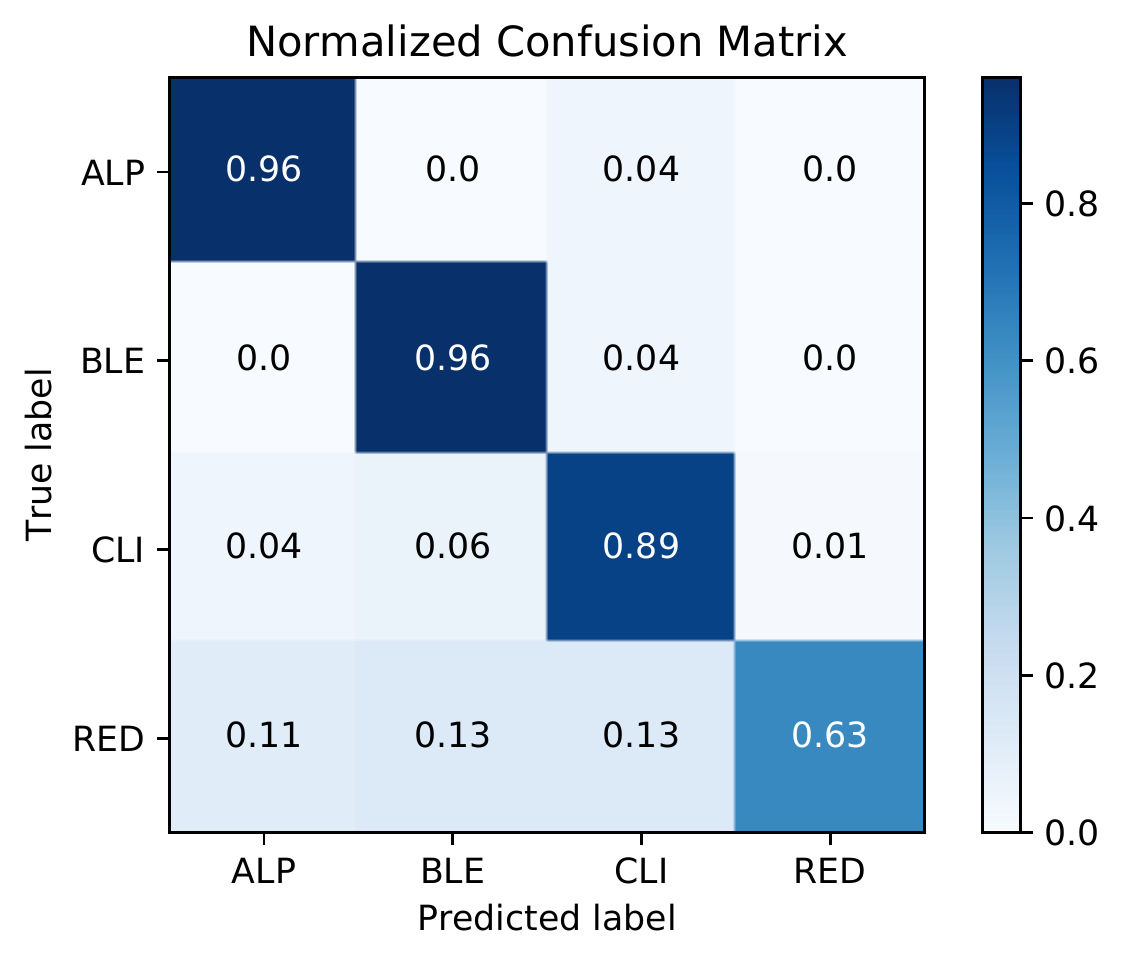}
		\caption{Confusion Matrix - 600BUD}
		\label{fig:morph_silver_standard}
	\end{subfigure}
	\caption{Performance of our character n-gram model on the gold and  600BUD test sets in closed class setting. Note good performance on all classes. (ALP: Alphabetisms BLE: Blends CLI: Clippings and REDUP: Reduplicatives).}
	\label{fig:morph_perf}
\end{figure}

\paragraph{Quantitative Evaluation on Gold Standard Test Set}
Table \ref{tab:f1_gold} shows the performance of the various models on the gold standard test set\footnote{Hyper-parameter tuning was done using cross validation.}.
Observe that both the morpheme and the character n-gram models substantially outperform the baseline \emph{random} classifier. 
Furthermore the model using character-ngram features significantly outperforms the morpheme based model. This is primarily because the morpheme based feature representation is too sparse for robust estimation of the decision boundary given the small amount of annotated training data. We believe that the morpheme features will be most effective when the amount of training data is much larger, to reduce the feature sparsity of general morpheme features.
Finally, Figure \ref{fig:morph_gold_standard} shows a confusion matrix for the character-ngram-based model when evaluated on the gold standard test set. 
It is evident that the classifier does well on \textsc{Alphabetisms, Clippings and Reduplicatives} but is relatively worse at identifying \textsc{Blends}.
We hypothesize this is because \textsc{Blends} are much more linguistically complex than other classes like \textsc{Alphabetisms} where distinctive features are much more easier to detect. 

In order to gain insight into the model that we learned using character n-grams, we examine the feature weights learned by the classifier. For  \textsc{Alphabetisms}, we find distinctive features which involve periods, upper-case letters and other symbols like {\&,/}.  For blends we observe strings that make up blends like \texttt{ny, sh} etc. Similarly for \textsc{Clippings} we observe distinctive suffixes like \texttt{ly-, ity, tie} while for \textsc{Reduplicatives} we see patterns like \texttt{-, ween, win, um} as well (used in words like \texttt{teenie-weenie} and \texttt{bum bum}) suggesting that our model effectively picks up on linguistic cues to effectively distinguish between the classes. 

\begin{table}[]
	\begin{tabular}{l|l}
	\textbf{Model} & \textbf{Weighted F1} \\
	\hline
	 Random & 26.57  \\
     Morpheme N-grams & 69.99 \\
     \textbf{Char N-grams} & \textbf{86.07} \\
	\hline
	\end{tabular}
	\caption{Performance of different models on Gold Standard Test set. Note that using character n-grams demonstrates the best performance.}
	\label{tab:f1_gold}
\end{table}

\paragraph{Inducing labels on the Urban Dictionary Data} As mentioned in Section \ref{sec:slang_classes}, slang does not exhaustively fall into the four categories we considered. Consequently, naively applying our learned model on \textsc{Urban Dictionary} data where slang can additionally belong to multiple ``unknown'' classes would result in poor performance (since several words which belong to none of the $4$ categories would be incorrectly assigned to one of these known categories). Observe that this is essentially the problem of \emph{open set recognition} studied quite rigorously in computer vision \cite{Scheirer_2014_TPAMIb,Scheirer_2017_TIFS,Jain_2014_ECCV,Rattani_2015_TIFS}, where several methods to address this problem are available. In our work, we consider one such approach that augments the trained model with a posterior probability estimator and a decision threshold to also optionally reject an instance.\footnote{We leave other complicated approaches which might boost performance on this task using W-SVM as future work.} Specifically, the approach consists of the following steps (see Algorithm \ref{alg:predict_reject}). Given $\mathcal{C}$, the set of known classes, and a closed set model $\mathcal{H}$ that outputs $Pr(y|\mathbf{X})$, a probability distribution over $\mathcal{C}$ given a feature vector $\mathbf{X}$, we want to make predictions using $\mathcal{H}$ on a data set $\mathcal{D}$ where instances could potentially belong to a unknown set of ``\textsc{unknown}'' classes in addition to $\mathcal{C}$. To do this, for each instance we compute a score signifying the confidence of $\mathcal{H}$ in its prediction and reject the instance if this confidence is below a manually chosen threshold $\delta$. This score could be (a) the maximum probability over the known classes or (b) the negative entropy of the output probability distribution. 

\begin{algorithm}[t!]
	\caption{\small \textsc{PredictWithReject} ($\mathcal{C}$, $\mathcal{H}$, $\mathcal{D}$, $\delta$, $\mathcal{F}$)}
	\label{alg:predict_reject}
	\begin{algorithmic}[1]
		\REQUIRE $\mathcal{C}$: Set of known classes, $\mathcal{H}$: Classifier for closed set $\mathcal{C}$, $\mathcal{D}$: Dataset of instances from an \emph{open set}. Each instance needs to be assigned a label in $\mathcal{C}$ or \textsc{Rejected} if it belongs to none of the classes in $\mathcal{C}$. $\delta$: Reject threshold. $\mathcal{T}$: Score type: \textsc{Negative Entropy} or \textsc{MaxProb}.   
		\ENSURE Predicted Label Assignments for each instance in $\mathcal{D}$ where each label $l\in\mathcal{C}\cup{\textsc{Rejected}}$
		\STATE \textsc{SCOREFUNC} $\gets \mathcal{F}$ \Comment {Initialize the scoring function to either compute the negative entropy or the maximum value of the output probability distribution.}
		\FOR{$e \in \mathcal{D}$}
		\STATE Compute $\mathbf{p}$ the output probability distribution over $\mathcal{C}$.
		\STATE $\textsc{Score}(w)$ $\gets$ \textsc{SCOREFUNC}$(w)$
		\STATE \textsc{Labels}(w) $\gets$ $\argmax_{c\in\mathcal{C}}\mathbf{p}$
		\IF {\textsc{SCORE}(w) $\le$ $\delta$}
			\STATE $\textsc{Labels}(w) \gets \textsc{Rejected}$ \Comment {Failed threshold so reject}
		\ENDIF
 		\ENDFOR  \\
 		\textbf{return} \textsc{Labels}
	\end{algorithmic}
\end{algorithm}
Table \ref{tab:top_examples_morpho} shows some of the top words detected by our model for each category on \textsc{UrbanDictionary} data. Our method effectively identifies instances of each class while also rejecting instances not belonging to four classes. We identify slang like \texttt{E.V.I.L and S.P.E.W} as \textsc{Alphabetisms} and detect \textsc{Blends} like \texttt{Iretalian:Irish+Italian} or \texttt{Obamerica:Obama+America}. Similarly our model is able to detect \textsc{Clippings} like \texttt{Stevie (Steven), Bishie (bishounen)} and \textsc{Reduplicatives} like \texttt{hooty-hoo}.

\begin{table}[]
	\begin{tabular}{l|p{6cm}}
		\hline
		\textbf{Category} & \textbf{Word} \\
		\hline
\textsc{Alphabetisms} & A.D.E.D, E.V.I.L, S.P.E.W, C.H.U.D, S.E.A.L, S.W.I.M \\
\hline	
		\textsc{Blends} & Iretalian (Ireland+Italian), Obamerica (Obama + America), Metapedia (Meta + Encyclopedia), Obroxious (Obnoxious+Rock), Cumbrella (Cum + Umbrella), Rainmelon (probably c.f No Rain + Blind Melon) \\
		\hline
		\textsc{Clippings} & cuttie (cutback), stevie (Steven), Bishie (bishounen), hattie, cottie \\ 
		\hline
		\textsc{Reduplicatives} & e-d-b-t-z, yu-gay-ho, yu-gay-oh, bug-a-boo, Roody-poo, hooty-hoo \\
		\hline
		\textsc{Rejected} & Darwin, edging, Pingo, Oil-Can, Wet-Seal, Flank, Baking \\
		\hline
	\end{tabular}
	\caption{Examples of top words detected by model for each class on the \textsc{UrbanDictionary} dataset along with words our algorithm correctly rejected. Observe that we can correctly identify labels for interesting words like \texttt{Ireitalian, obroxious and cuttie}.}
	\label{tab:top_examples_morpho}
\end{table}

We also evaluate our model quantitatively in a closed class setting on a balanced manually created test sample of the \textsc{UrbanDictionary} data-set of $600$ words (\textbf{600BUD}) over which we obtained a weighted F1-score of $86\%$ (see Figure \ref{fig:morph_silver_standard} for the confusion matrix on \textbf{600BUD}). Finally, we evaluated our model in the open class setting using \emph{cross-class validation}\cite{Scheirer_2014_TPAMIb} which yields a mean weighted F1 score of $66.43$ implying that our model generalizes reasonably well to this open set recognition setting as well.

\subsection{Syntax}
\textbf{How do the syntactic roles in which slang words are used differ from those in Standard English?}
We investigate this by analyzing the part of speech (POS) roles of slang. We obtain the POS tags for slang words by using a pre-trained part of speech tagger\footnote{We use TextBlob for inferring POS tags.} on the example usages of slang. We also obtained the POS tags of words in Standard English by querying \emph{Dictionary.com}. We observed that slang contains a much higher proportion of \textsc{Nouns} ($\sim 72\%$). In contrast, the fraction of \textsc{Nouns} in Standard English was $\sim 50\%$. Further analysis reveals that about $28\%$ of slang are used as \textsc{Proper Nouns}. We explain this by noting that even names of people can be used as slang (with a creatively assigned connotation, as we will see in Section \ref{sec:social}) which is quite rare in the standard form. Examples of such words include \texttt{Trumpence, Angry Bill Cosby, Erik Erikson, Annabelle, Ria, Debby} etc.

%

\section{Social Aspects of Slang Usage}
\label{sec:social}
According to the sociological viewpoint of defining slang \cite{mattiello2008introduction}, slang is associated with several sociological properties with perhaps the most widely accepted one being \emph{group restriction}. In addition to this, slang also exhibits properties like \emph{debasement, humor, obscenity, and subject-restriction} to name a few. While several scholars have studied the social aspects of slang they have been largely qualitative \cite{drake1980social} or restricted to a very specific group \cite{bucholtz2006word}, a quantitative large scale analysis of social aspects of Internet slang has not been addressed to the best of our knowledge. Here, we consider two such aspects: (a) \textbf{Subject restriction}: Slang can be associated with a particular subject. Examples of such slang are \texttt{crack, junkie, acid, crystal}, all related to the subject of \textsc{drugs}. Similarly, slang words abound in obscenity with a plethora of terms related to \textsc{Sex}. (b) \textbf{Stereotypes and Prejudices}: We investigate the question of whether slang like the standard form  reflects prevalent gender and religious biases/prejudices.  
\subsection{Subject Restriction in Internet Slang}
\begin{figure}[]
\begin{subfigure}{0.475\columnwidth}
	\includegraphics[width=\columnwidth]{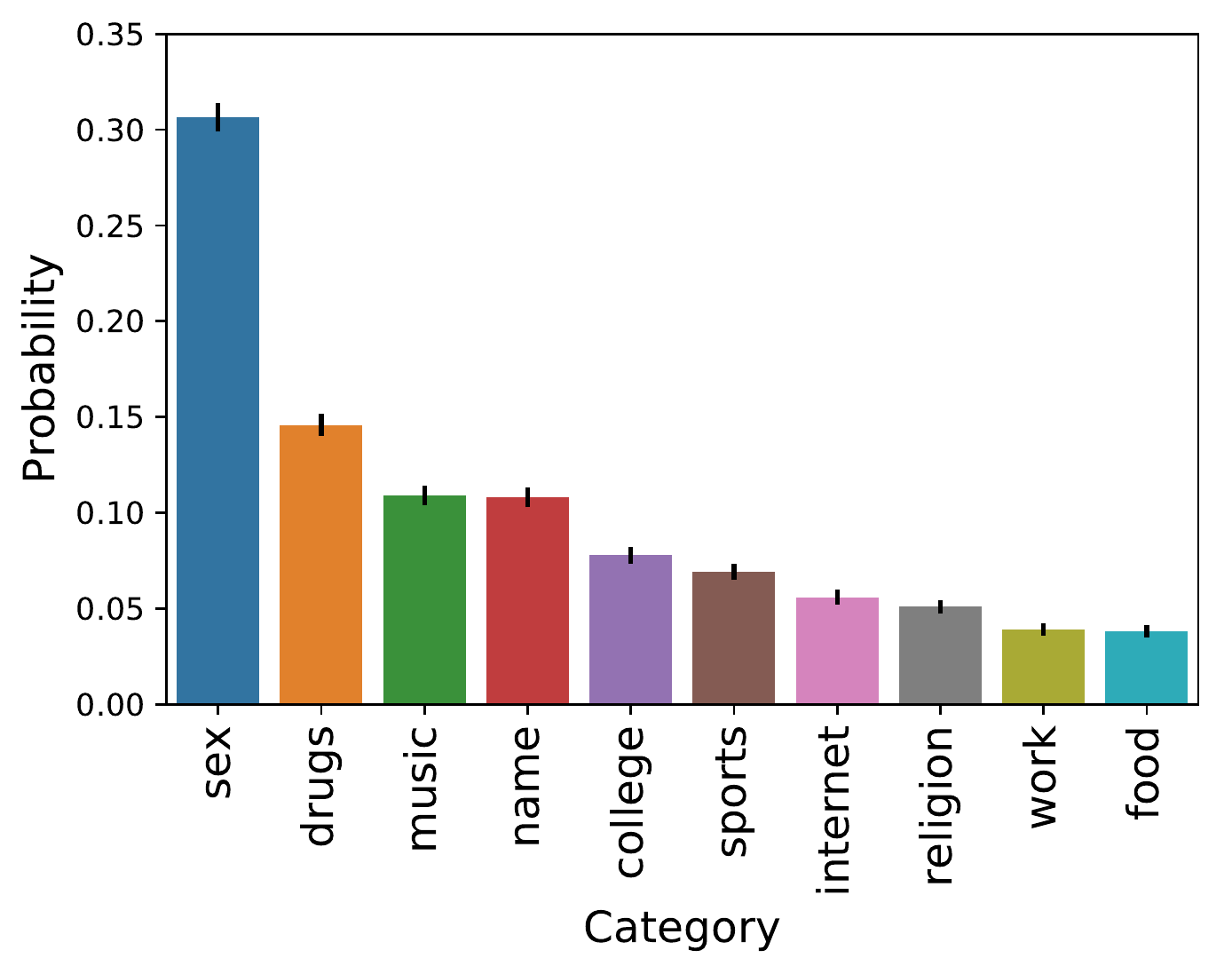}
	\caption{Gold Standard Data}
	\label{fig:words_categ_prob_gold}
\end{subfigure}
\begin{subfigure}{0.475\columnwidth}
	\includegraphics[width=\columnwidth]{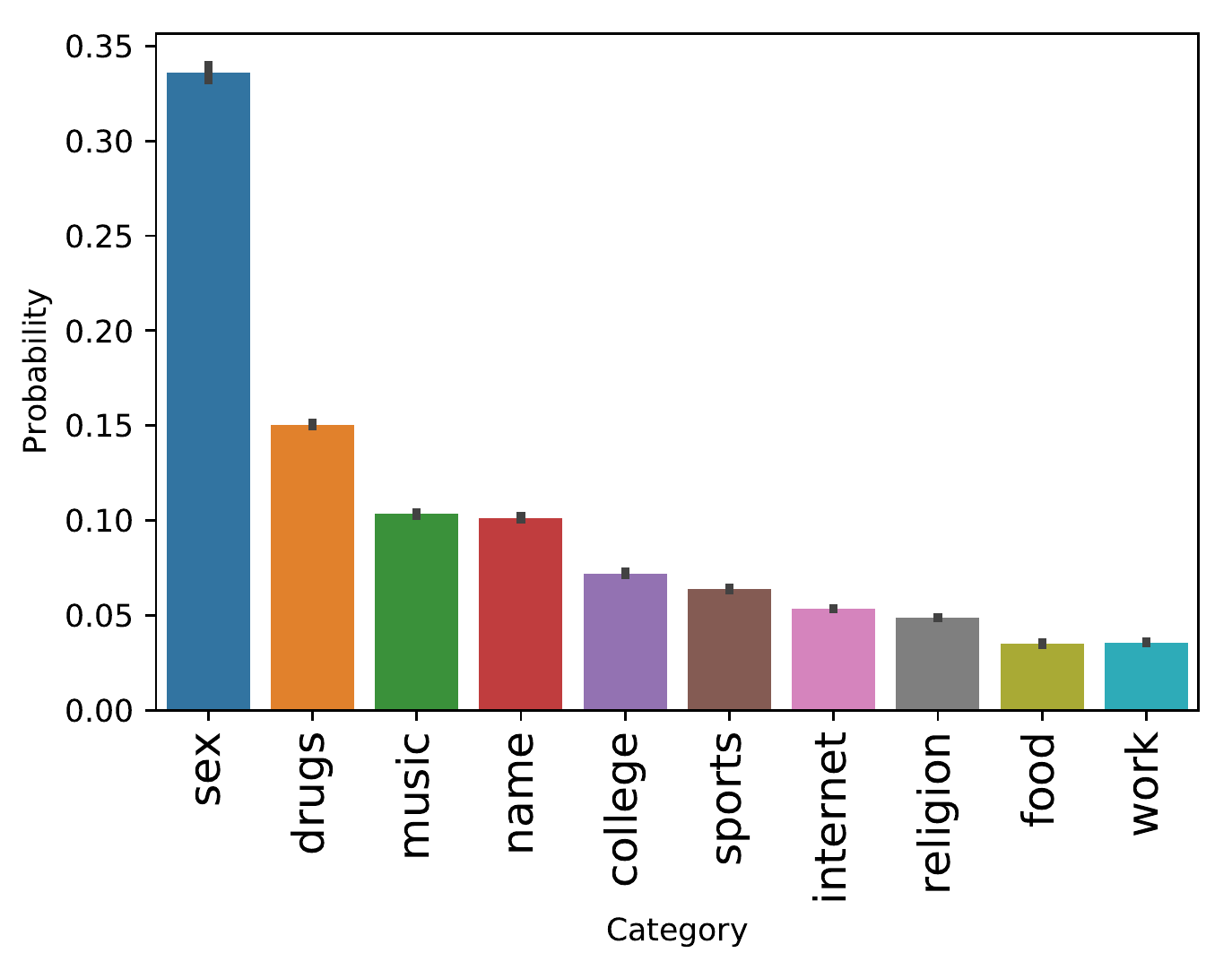}
	\caption{Gold Standard Data + 11K}
	\label{fig:words_categ_prob_all}
\end{subfigure}
\caption{Subject restriction in slang. Note the dominant prevalence of \textsc{Sex} and \textsc{Drugs} reinforcing the obscene and vulgar aspects often associated with slang. Note, the estimated proportion of \textsc{Sex} is even higher when we include the additional 11K examples.}
\label{fig:words_categ_prop}
\end{figure}

\textsc{UrbanDictionary} categorizes slang words into one of $10$ categories when possible, namely: \textsc{Sex} (eg. bating, spear), \textsc{Drugs} (eg. weeded, blower), \textsc{Music} (eg. bridestep, tweenwave), \textsc{Name} (eg. bati, hannah-montana), \textsc{College} (eg. architorture), \textsc{Sports} (huck, poned), \textsc{Internet} (eg. typeractive, eracism), \textsc{Religion} (eg. kyke, jooz),\textsc{ Food} (eg. grubbin, scram) and \textsc{Work} (eg. pixel-counting, vandy). 
It is to be noted that these categories are not exhaustive and once again form an open-set (there are are many slang words that belong to multiple "unknown" classes in addition to the $10$ known classes). Figure \ref{fig:words_categ_prob_gold} shows the fraction of each category using this gold standard labeled data. Observe that the top $2$  categories are \textsc{Sex} and \textsc{Drugs} suggesting the dominance of these topics in slang.

Given a small number of gold standard annotated data from \textsc{UrbanDictionary}, we seek to infer labels for the much larger un-annotated data-set noting that the labels are not exhaustive. As before, we learn a classifier on the supervised labeled data and extend it to handle open-set recognition using Algorithm \ref{alg:predict_reject}. 

\subsubsection{\textbf{Learning Slang Embeddings}}
To capture semantics of slang words, we learn embeddings of slang words by capturing distributional cues based on their usage in examples.  To illustrate, one example usage of \texttt{thizz} is \emph{``"thizz is NOT pure extacy.....thizz is a bunch of shitty drugs ( meth, heroin, extacy, coke, and acid if 
your lucky...;)  ) all mixed up and put into a pill press''}, which suggests that \texttt{thizz} is a slang that is related to \textsc{drugs}. We learn Skipgram \cite{mikolov2013linguistic} word embeddings of dimension $d=100$ using negative sampling.

\subsubsection{\textbf{Learning the Categorization Model}}
Using the slang embeddings learned as features, we learn a K-nearest neighbor classifier \footnote{Other classifiers like SVM yield comparable results.} on the gold standard annotated dataset (we set $k$=5).
To quantitatively evaluate the performance of our learned model, we manually created a labeled test set of $200$ words. Our model obtained a weighted F1-score of $63.22\%$ on this test set. In contrast, predicting a label according to the training data label distribution yields an F1-score of $13.45\%$. We further observed that our classifier does the best on words belonging to \textsc{Sex}, \textsc{Drugs}, \textsc{Internet}, and \textsc{Music} and \textsc{Sex} which are more prominent and performs quite poorly on \textsc{Food} and \textsc{Religion} which occur infrequently in slang.

Finally, we extend our model to the "open-set" recognition setting by augmenting the learned classifier with a probability threshold, using Algorithm \ref{alg:predict_reject} and apply our model on an additional $11000$ (\textbf{11K}) unlabeled slang words. 
Table \ref{tab:top_categ} shows the top words identified by our learned model for $5$ categories as exemplars. Specifically, note that words like \texttt{pipe, shots, herb, roll, cigs} are correctly classified by our model as being related to \textsc{Drugs}. More generally, observe that most of the words in each category are representative of their corresponding class, suggesting that our learned model is able to effectively pick up on contextual usage of the word to learn its categorization. Finally using the labels inferred by our model on these $11000$ words and including the gold standard dataset we created a larger dataset. We estimated the mean proportion of each subject (at different rejection thresholds) on this larger dataset which clearly suggests that \textsc{Sex} and \textsc{Drugs} are dominant topics in slang (see Figure \ref{fig:words_categ_prob_all}) consistent with our previous observation.

\begin{table}[htb!]
	\begin{tabular}{l|p{6cm}}
		\hline
		\textbf{Category} & \textbf{Word} \\ 
		\hline
		\textsc{Sex} & fine ass, bum hole, penis, foreskin, entrails, tooth, finger, hard dick, facial hair \\
		\hline	
		\textsc{Drugs} & pipe, shots, herb, dwayne, dead presidents, nuggs, bomb, roll, cubba, cigs \\
		\hline
		\textsc{College} & med school, quiz, league, good boy, math, dropout, derbie, preppy, herd \\
		\hline 
        \textsc{Food} & pepperoni, chocolate chip, ice cream, tortilla, cake, chili, hot dog, gluten \\
        \hline
       \textsc{Rejected} & billion, copy, dragons, drone, terrorism \\
       \hline
	\end{tabular}
	\caption{Examples of Top words in 4 of the 10 categories detected by model with Algorithm \ref{alg:predict_reject} along with examples of words correctly rejected. Note how words like \texttt{pipe, shots and herb} are correctly categorized as  belonging to \textsc{Drugs} while sexual terms like \texttt{entrails} are correctly identified as belonging to \textsc{Sex}.}
	\label{tab:top_categ}
\end{table}

\subsection{Stereotypes in Internet Slang}
Recent research has shown that word embeddings learned on data (where language is standard) typically  from \textsc{Wikipedia} and even the more formal \textsc{Google News} reveals gender biases and reinforces existing gender stereotypes \cite{bolukbasi2016man,caliskan2016semantics,zhao2017men}. Inline with these observations we investigate this in slang through the lens of slang embeddings.
\textbf{Does slang reflect gender stereotypes?}
While traditional gender stereotypes with respect to occupations are prevalent in News and even on Wikipedia, one might posit that since the user demographics of UrbanDictionary is skewed towards younger age groups where $18-24$ and $25-34$ are significantly overepresented \footnote{Data from Amazon Alexa.}, prevalent occupational stereotypes with respect to gender might be weaker than what one might observe in general language.
    
\begin{figure}[]
	\includegraphics[width=\columnwidth]{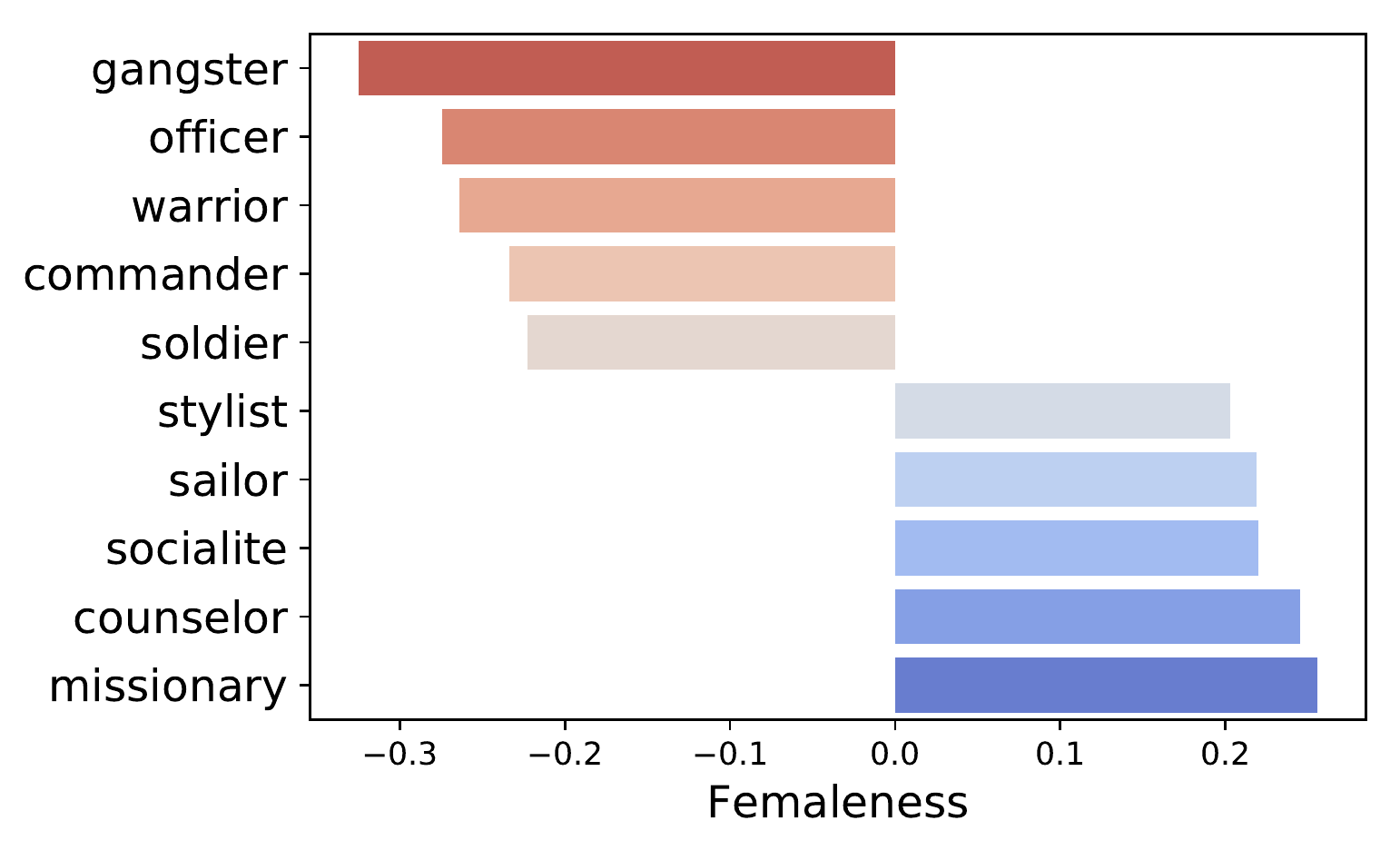}
	\caption{Gender Stereotypes for occupations in Slang. Note the high ``maleness'' associated with occupations like \texttt{gangster, officer} and the high ``femaleness'' associated with occupations like \texttt{socialite, counselor, missionary}.}
	\label{fig:gender_stereotypes}
\end{figure}

To answer this question, we follow the method outlined by \cite{bolukbasi2016man} to quantify gender bias and stereotypes using their pre-defined humanly validated list of occupations.  We applied their method to quantify \emph{direct bias} (\textsc{DirectBias}$_{1}$) \cite{bolukbasi2016man} on the slang embeddings.
Ideally, if there is no gender bias, this would be $0.0$. However, we notice a significant \textsc{DirectBias}$_{1}=0.09$ for slang embeddings (corresponding value for \textsc{GoogleNews} was $0.08$ \cite{bolukbasi2016man}). This suggests that slang also reflects traditional gender stereotypes at a level comparable to other standard language used in News. Finally, Figure \ref{fig:gender_stereotypes} shows the extreme $5$ occupations based on their projections onto the gender axis using slang embeddings. Note that the occupations \emph{gangster, officer, warrior, commander, soldier} are considered much more \textsc{Male} whereas occupations like \emph{stylist, sailor, socialite, counselor and missionary} are considered to be much more \textsc{female} reflecting  and reinforcing prevalent gender stereotypes.

\textbf{Conclusion}: Similar to standard language and even formal news, slang is not immune to gender biases and stereotypes. 

\textbf{Does slang exhibit sexual prejudice?}
Noting the significant proportion of sexual terms in slang, it is natural to ask the question \emph{Does slang exhibit sexual prejudices when referring to males/females?}

\begin{figure}[t!]
	\includegraphics[width=\columnwidth]{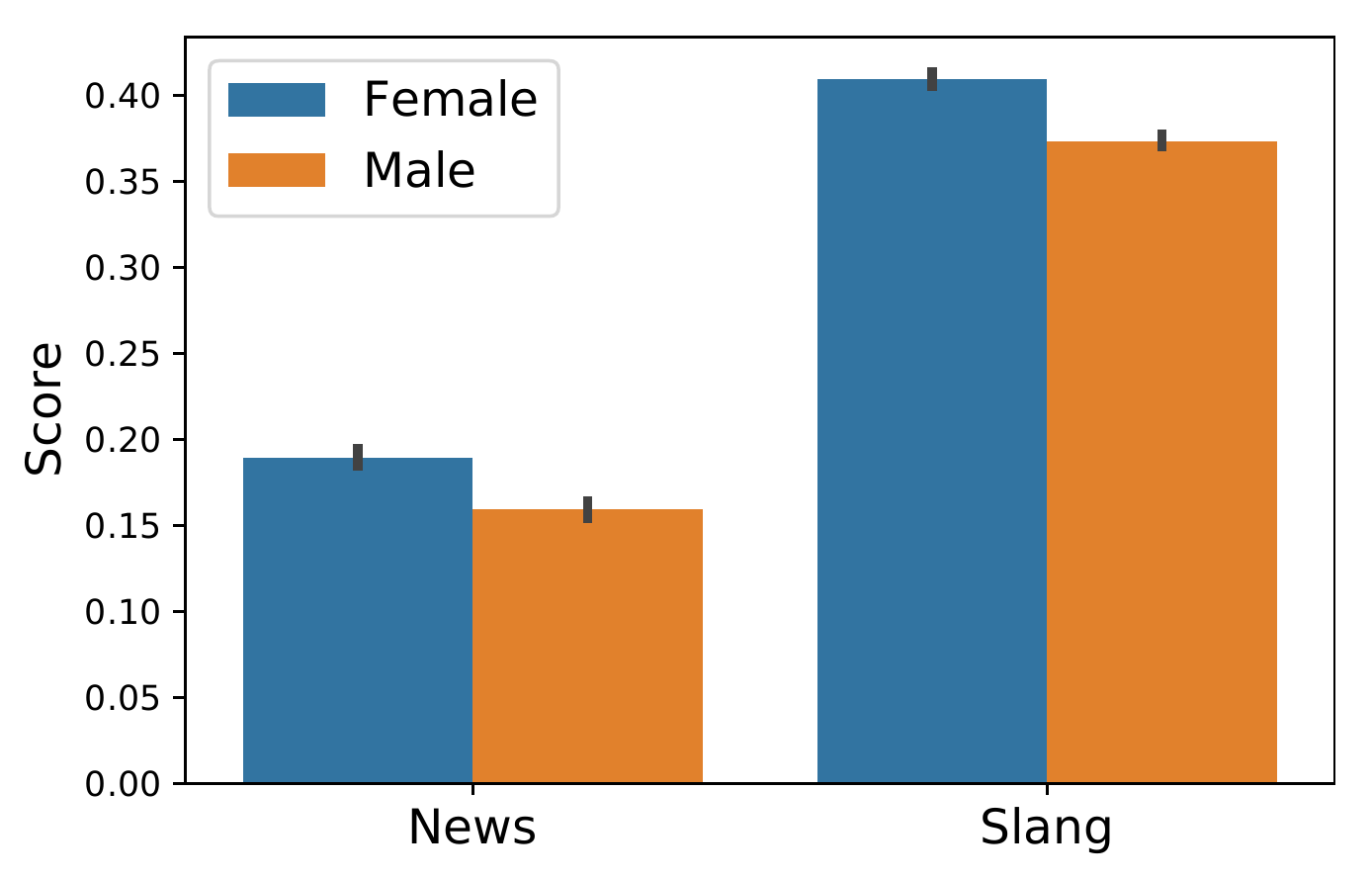}
	\caption{Sexual Prejudice of person names. Note that both male and female names demonstrate higher sexual prejudice in slang than a more formal variety (NEWS). Furthermore, female names are more sexually prejudiced than male names in both varieties.}
	\label{fig:sexual_prejudices}
\end{figure}

To answer this question, we first observe that several person names are also added as slang entries (for example. female names like \texttt{neelam, sangeeta, ganga, ria, annabelle} and male names like \texttt{vance, reuben}). First, we obtain a pre-compiled list of words (not an exhaustive one) primarily associated with sexual prejudice namely \texttt{slut, whore, shrew, bitch, faggot, sexy, fuck, fucked, nude, porn, cocksucker}\footnote{Obtained from: https://github.com/commonsense/conceptnet5/}. Given a word $w$, we define its sexual prejudice score as the mean cosine similarity of the word embedding for $w$ and each of the words in the above set. Specifically, we define:
$SEXPREJ(w)=\frac{\sum_{c \in L}Cosine(w,c)}{|L|}$. We then obtained a list of $\sim5000$ slang words which are names of persons, inferred their gender (male or female) using a pre-trained model and computed their sexual prejudice score using slang embeddings. \footnote{We use the \texttt{sexmachine} toolkit to infer gender.}
As a baseline, we also computed the corresponding sexual prejudice scores for the same set of names using the pre-trained \textsc{GoogleNews} word embeddings.
Figure \ref{fig:sexual_prejudices} shows the mean sexual prejudice score for males and females in both Slang and Google News. 
Observe that in both Google News and Slang, both male and female names are associated with sexual prejudices. Furthermore, sexual prejudices associated with female names is (statistically significantly) higher than male names in both sources. More interestingly, we observe that sexual prejudice associated with people names is significantly higher in Slang than in a more standard and formal domain like News. Some examples of female names which reflect such sexual prejudices in slang from \textsc{UrbanDictionary} which would be very unlikely in formal domains like News are (a) \texttt{ria}:\emph{hey tht chic is such a ria', u'hey look...its ria!.} and (b) \texttt{debby}:\emph{That girl is such a debby, the guys are always checking her out}.

\textbf{Conclusion}: Our analysis suggests that slang is much more uninhibited with stronger sexual prejudices. 

\begin{figure*}[ht!]
	\begin{subfigure}{0.24\linewidth}
		\includegraphics[width=\columnwidth]{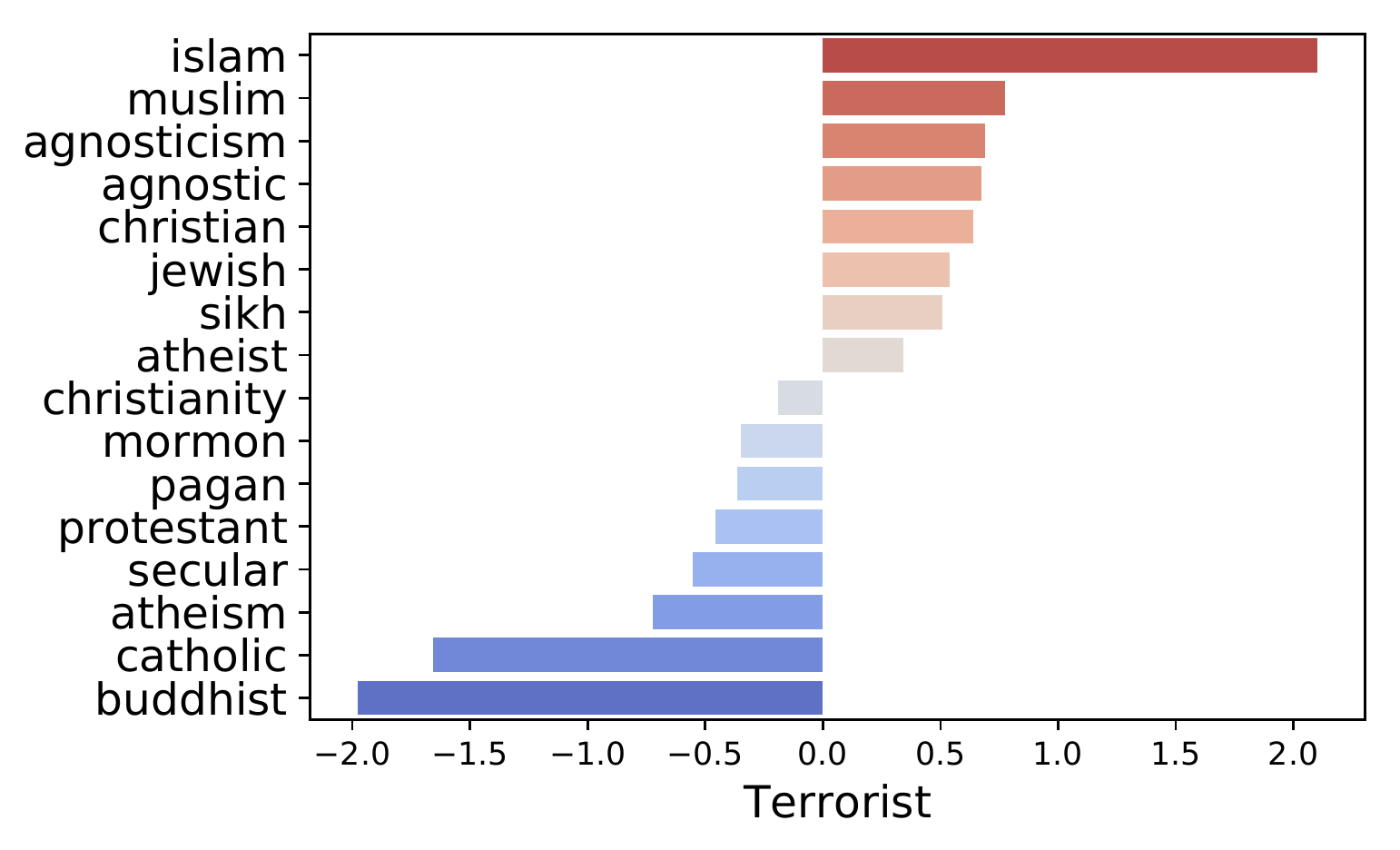}
		\caption{Terrorist}
		\label{fig:A}
	\end{subfigure}
	\begin{subfigure}{0.24\linewidth}
		\includegraphics[width=\columnwidth]{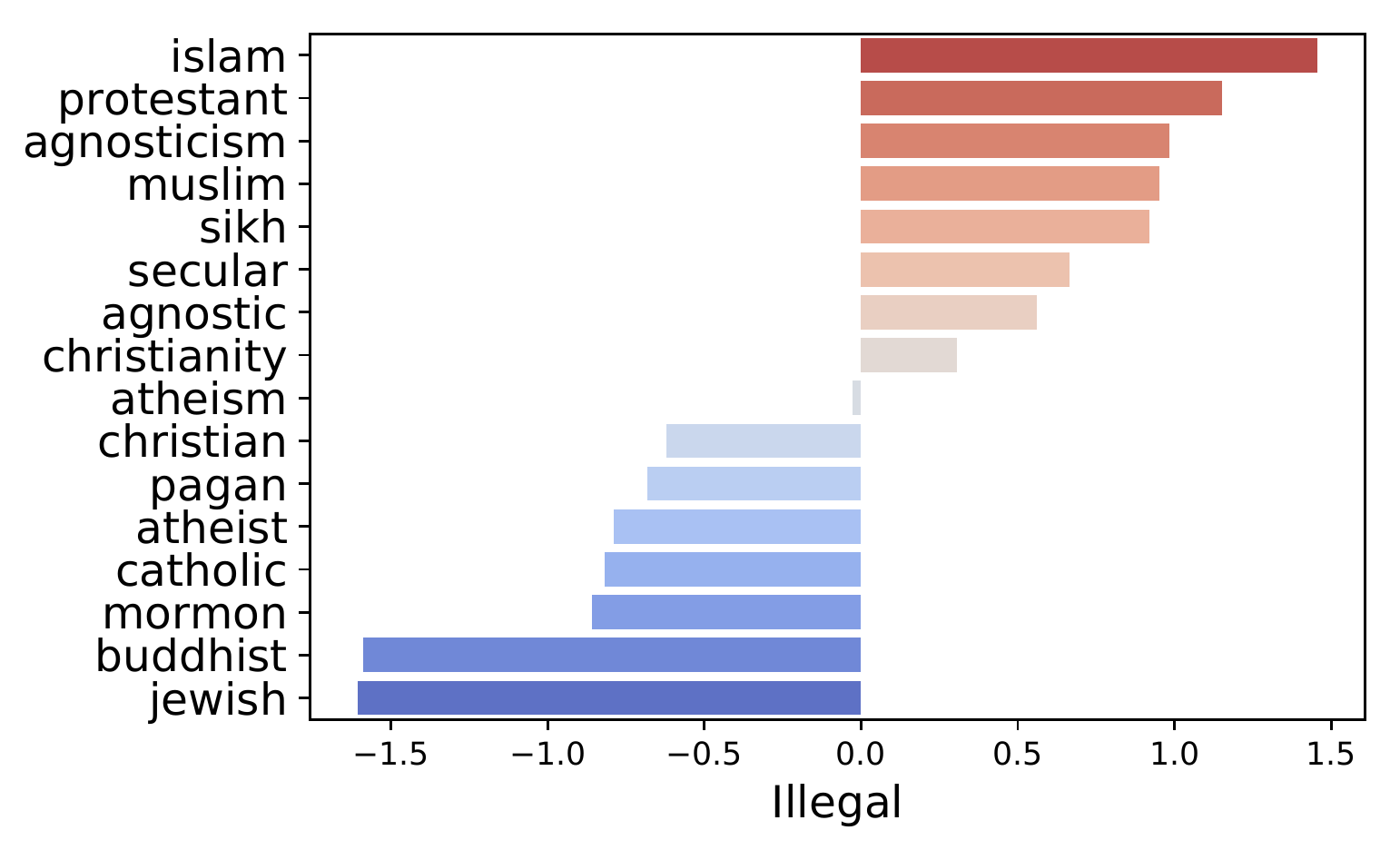}
		\caption{Illegal}
		\label{fig:B}
	\end{subfigure}	
	\begin{subfigure}{0.24\linewidth}
		\includegraphics[width=\columnwidth]{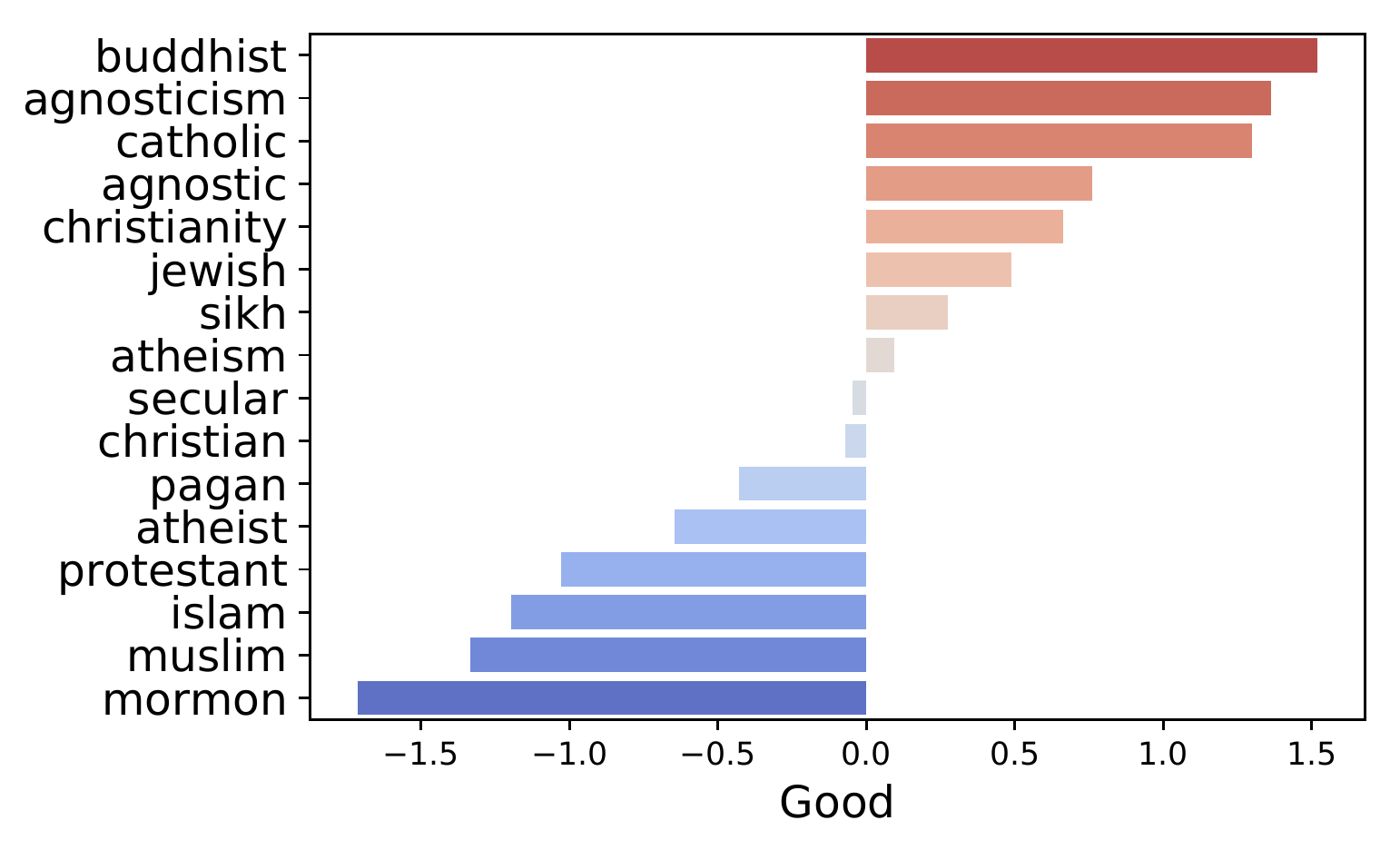}
		\caption{Good}
		\label{fig:D}
	\end{subfigure}
	\begin{subfigure}{0.24\linewidth}
		\includegraphics[width=\columnwidth]{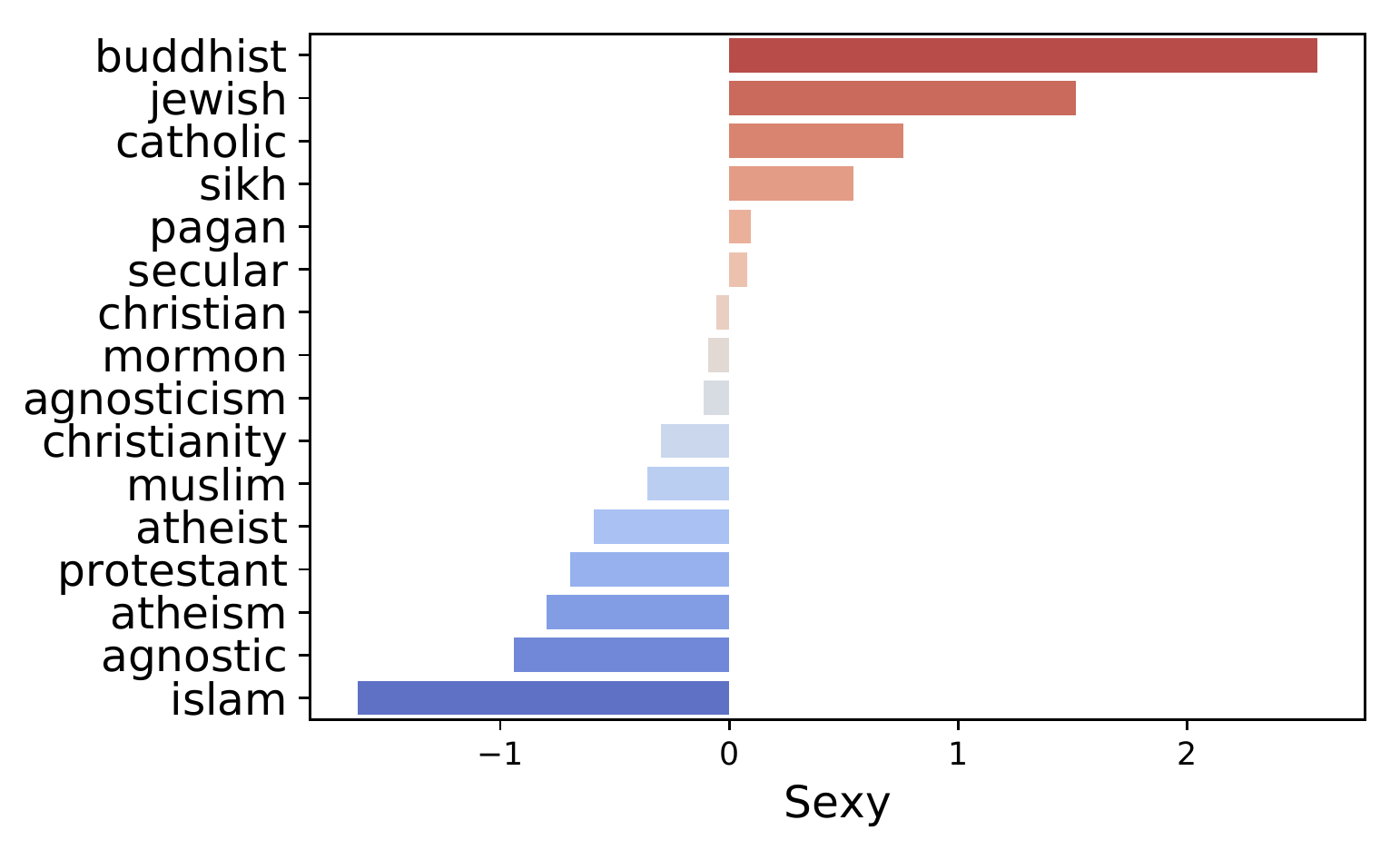}
		\caption{Sexy}
		\label{fig:E}
	\end{subfigure}
	\caption{Religious prejudices in Slang. Note that even in slang, prevalent religious prejudices manifest. For example: \texttt{Islam, Muslim} are closer to \texttt{terrorist} (see Figure \ref{fig:A}). Note \texttt{atheist} has near neutral prejudice score while words like \texttt{Buddhist} have negative prejudice scores. Similarly observe the reversal of religions in positive trait words like \texttt{good, sexy}.} 
	\label{fig:religion}
\end{figure*}

\textbf{Does slang show religious prejudices?}
We now investigate the prevalence of religious prejudices in slang. First, we define two lists:
(a) \textbf{Religious terms} a list of religious terms spanning multiple religions. These include words like \texttt{sikh, muslim, islam, jews, christian, agnostic, atheist, agnostic, buddhist}.
(b) \textbf{Prejudice terms} which reflect prejudices. Examples of such words include \texttt{terrorist, evil, sexy, suave and good}. These lists are in no-way exhaustive but representative. \footnote{We obtained these lists from: https://github.com/commonsense/conceptnet5} 
Given a religious term $r$ and a prejudice term $p$ we compute a prejudice score for $(r,p)$ defined as the cosine similarity of $r$ with $p$. We standardize these scores over all religious terms for a given prejudice $p$. Figure \ref{fig:religion} shows the standardized scores obtained for each of the religious terms for several prejudice terms. Note how for prejudices \texttt{terrorist, and illegal}, religious terms \texttt{Islam, muslim, and christian} have the highest scores where as \texttt{buddhist and jewish} have the most negative scores (see Figures \ref{fig:A}, \ref{fig:B}). Furthermore observe that for positive traits \texttt{good, sexy} (Figures \ref{fig:D}, \ref{fig:E}), \texttt{buddhist, jewish} are associated with very high positive scores while \texttt{islam, agnostic, muslim} are associated with extreme negative scores thus reinforcing existing stereotypes. Finally we observed the mean prejudice score over all $(r,p)$ pairs in slang was significantly greater than \textsc{Google News} ($0.34$ vs $0.16$ pval<0.0001) suggesting that such prejudices manifest more extremely in slang. \textbf{Conclusion}: We show that slang usage reveals prevalent religious stereotypes suggesting that slang like standard language is subject to the same biases and stereotypes prevalent in standard language.


\section{Related Work}
\label{sec:related}
\textbf{Socio-variational Linguistics} A large body of work studies linguistic aspects of language and its correlation with social factors like age, gender, ethnicity, and geography\cite{mencken1945american,romaine1983locating,ruin,eisenstein2010latent,eisenstein2011discovering,eisenstein2012mapping,bamman2014gender}. Most of these works either study the standard form of English (in written or online social media) and do not focus primarily on slang. \citet{mencken1945american} outlines variation between American English and British English and \citet{romaine1983locating} studied language variation in time and geography,and outlined principles of language change. In the age of social media, \citet{eisenstein2010latent,eisenstein2011discovering,eisenstein2012mapping} study lexical variation in social media, propose models to detect geographic lexical variation in social media and study its diffusion across regions. \citet{bamman2014gender} then follow-up by studying gender identity and lexical variation in social media. 

There has been little work on the linguistic and social aspects of slang with the exception being the work of \citet{mencken1967american} who studies the origin and nature of American Slang. Consequently, few dictionaries documenting slang have been compiled before the evolution of Internet and social media \cite{flexner1967dictionary,chapman1986new}.  Recently, \citet{mattiello2008introduction} notes and provides qualitative evidence for the extragrammatical morphological properties of slang while some works attempt to explicitly incorporate slang to improve tasks in natural language processing (like sentiment detection) especially for social media like Twitter \cite{nielsen2011new,amiri2012mining,kundi2014detection,dhuliawala2016slangnet}. 

 The works that are closest to ours are that of  \citet{mattiello2005pervasiveness,mattiello2008introduction,dhuliawala2016slangnet}. \citet{dhuliawala2016slangnet} use Reddit and \textsc{UrbanDictionary} to build a lexical resource called \textsc{SlangNet} that captures slang semantics.
\citet{mattiello2005pervasiveness, mattiello2008introduction} notes the pervasiveness of slang on the Internet, outlines the extra-grammatical nature of slang morphology and compiles a small dataset of $1580$ slang words.
 Differing from all of these works, we distinguish ourselves by explicitly focusing and analyzing slang on both aspects: linguistic and social at a large scale. We not only provide supporting quantitative evidence for observations made by \citet{mattiello2008introduction} but additionally conduct the first phonological, morphological and syntactical analysis of tens of thousands of slang words -- revealing new linguistic insights into diverse patterns of slang generation.

\textbf{Fairness in Machine Learning} There has been a surge of research into quantifying bias and analyzing fairness of machine learning models including word embeddings \cite{hajian2016algorithmic,hardt2016equality,zliobaite2015survey,bolukbasi2016man,bolukbasi2016quantifying,crawford2016artificial}. Among these, the  most relevant works are by \cite{bolukbasi2016man,bolukbasi2016quantifying} who analyze and quantify  the gender bias prevalent in pre-trained word embedding models like \textsc{Google News} word embeddings. They also propose methods to debias such word embeddings. Our work builds on their approach where we not only quantify such bias in slang, but also reveal the presence of additional sexual and religious prejudices.   
\section{Conclusion}
\label{sec:Conclusions}
In this work, we conducted the first large scale analysis of slang on the Internet on both aspects: linguistic and social. Our linguistic analysis of slang, which included phonological, morphological and syntactical analysis reveals that slang exhibits linguistic properties markedly different from the standard variety. Furthermore our analysis reveals insights into generative mechanisms of four different classes of slang: \textsc{Alphabetisms}, \textsc{Blends}, \textsc{Clippings} and \textsc{Reduplicatives}. We also propose a model to classify slang words into these categories yet effectively \emph{reject} words which do not belong to any of these four known classes.

We also analyzed social aspects of slang pertaining to subject restriction and stereotyping. Our analysis revealed two dominant subjects of slang: \textsc{Sex} and \textsc{Drugs}. Additionally, we showed that slang like its standard variety exhibits a non-trivial gender bias. More interestingly, our analysis reveals that both male and female names are disproportionately sexualized in slang. In general, we noted that slang is not immune to prevalent sexual and religious prejudices and in-fact manifests such prejudices to a greater degree.  

Our work also suggests several directions for future research. First, our analysis is restricted to slang in UrbanDictionary which mostly reflects slang usage on the Internet and as used by youth. It would be interesting to analyze slang in alternate settings like forums, micro-blogs, printed media, audio/video content as well as specific demographic groups (like senior citizens etc). Second, our insights can enable development of generative models for slang and its diffusion in social media. Thirdly, the methods outlined in our work can enable a study of slang in the multi-lingual and cross-cultural setting where slang usage and its social aspects can be quite culture specific. 

Finally, we conclude by noting that our work has implications to the larger fields of Internet Linguistics and natural language processing which is increasingly being applied to non-standard language varieties so prevalent in social media. 
\bibliographystyle{ACM-Reference-Format}
\bibliography{sample-bibliography} 

\end{document}